\documentclass[journal]{IEEEtran}
\usepackage{amsmath,amsfonts}
\usepackage{array}
\usepackage{booktabs}
\usepackage{multirow}
\usepackage{textcomp}
\usepackage[table]{xcolor}
\usepackage{graphicx}
\usepackage{cite}
\usepackage{url}
\usepackage[switch]{lineno}
\usepackage{subcaption}
\usepackage{tcolorbox}
\definecolor{headergray}{gray}{0.9}
\newcommand{\impgain}[1]{\textsubscript{$\uparrow$#1}}
\newcommand{\venuecite}[2]{\textcolor{gray}{\textsubscript{[#1]}}~\cite{#2}}
\begin{document}
	
	\title{Q-Zoom: Query-Aware Adaptive Perception for Efficient Multimodal Large Language Models}
	
	\author{Yuheng Shi, Xiaohuan Pei, Linfeng Wen, Minjing Dong, Chang Xu
		\thanks{Yuheng Shi (yshi0087@uni.sydney.edu.au), Xiaohuan Pei (xiaohuan.pei@sydney.edu.au), and Chang Xu (corresponding author, c.xu@sydney.edu.au) are with University of Sydney, Australia.}%
		\thanks{Linfeng Wen (wenlf5@mail2.sysu.edu.cn) is with Sun Yat-sen University, China.}%
		\thanks{Minjing Dong (minjdong@cityu.edu.hk) is with City University of Hong Kong, China.}%
	} 
	
	\markboth{Preprint}
	{Anonymous Author(s): Your TPAMI Paper Title}
	\maketitle

	\begin{abstract}
	Multimodal Large Language Models (MLLMs) require high-resolution visual inputs for fine-grained tasks like document understanding and dense scene perception. However, current global resolution scaling paradigms indiscriminately flood the quadratic self-attention mechanism with visually redundant tokens, severely bottlenecking inference throughput while ignoring spatial sparsity and query intent. To overcome this, we propose \textbf{Q-Zoom}, a query-aware adaptive high-resolution perception framework that operates in an efficient coarse-to-fine manner. First, a lightweight Dynamic Gating Network safely bypasses high-resolution processing when coarse global features suffice. Second, for queries demanding fine-grained perception, a Self-Distilled Region Proposal Network (SD-RPN) precisely localizes the task-relevant Region-of-Interest (RoI) directly from intermediate feature spaces. 
	To optimize these modules efficiently, the gating network uses a consistency-aware generation strategy to derive deterministic routing labels, while the SD-RPN employs a fully self-supervised distillation paradigm. A continuous spatio-temporal alignment scheme and targeted fine-tuning then seamlessly fuse the dense local RoI with the coarse global layout.
	Extensive experiments demonstrate that Q-Zoom establishes a dominant Pareto frontier. Using Qwen2.5-VL-7B as a primary testbed, Q-Zoom accelerates inference by $2.52\times$ on Document \& OCR benchmarks and $4.39\times$ in High-Resolution scenarios while matching the baseline's peak accuracy. Furthermore, when configured for maximum perceptual fidelity, Q-Zoom surpasses the baseline's peak performance by 1.1\% and 8.1\% on these respective benchmarks. These robust improvements transfer seamlessly to Qwen3-VL, LLaVA, and emerging RL-based thinking-with-image models, setting a new state-of-the-art for efficient, fine-grained visual perception. Project page is available at \url{https://yuhengsss.github.io/Q-Zoom/}.
\end{abstract}


	\section{Introduction}
\IEEEPARstart{M}{ultimodal} Large Language Models (MLLMs)~\cite{liu2023llava,Qwen2.5-VL,zhu2025internvl3} have rapidly emerged as the cornerstone of artificial general intelligence, demonstrating unprecedented capabilities in visual reasoning~\cite{zheng2025deepeyes,zhang2025thyme,lai2025mini}, document understanding~\cite{li2024monkey,huang2024mini,wei2025deepseek}, and vision-language-action (VLA) modeling~\cite{driess2023palm, kim2024openvla, black2024pi_0}. The bedrock of these sophisticated reasoning capabilities lies in the model's foundational visual perception. 
Early pioneer architectures~\cite{dai2023instructblip,liu2023llava,liu2023improvedllava,liu2024llavanext}, such as the original LLaVA series~\cite{liu2023llava,liu2023improvedllava}, relied on frozen Vision Transformers~\cite{dosovitskiy2020image,tschannen2025siglip,radford2021learning} operating at a low and fixed resolution. While effective for coarse-grained image captioning, this rigid paradigm severely compressed and blurred critical local details. To overcome this perceptual bottleneck, subsequent works~\cite{zhang2024llava,liu2024llavanext, Qwen2-VL, chen2024internvl} have significantly advanced high-resolution adaptation. Sophisticated architectures, such as the AnyRes strategy~\cite{chen2024internvl,liu2023improvedllava} or native dynamic resolution encoding~\cite{Qwen2-VL,dehghani2023patch}, allow models to ingest varying and significantly higher input resolutions. 
These methods have achieved remarkable leaps in fine-grained perception and have become foundational mechanisms in recent state-of-the-art (SOTA) MLLMs~\cite{zhu2025internvl3, Qwen2.5-VL}.

Despite these advancements, efficient visual perception remains challenging. Current dynamic resolution solutions default to a brute-force scaling paradigm, producing visual tokens based solely on raw input resolution. Because visual representations serve to answer specific user queries, this exhaustive approach is computationally prohibitive and redundant. First, it ignores \textit{query-level intent} by assuming maximum visual fidelity is universally required, wasting resources on simple questions answerable with coarse features. Second, it ignores \textit{spatial sparsity}. Globally scaling the entire image floods the LLM's quadratic self-attention mechanism with thousands of visually useless background tokens.

\begin{figure*}[t]
	\centering
	\includegraphics[width=1.0\linewidth]{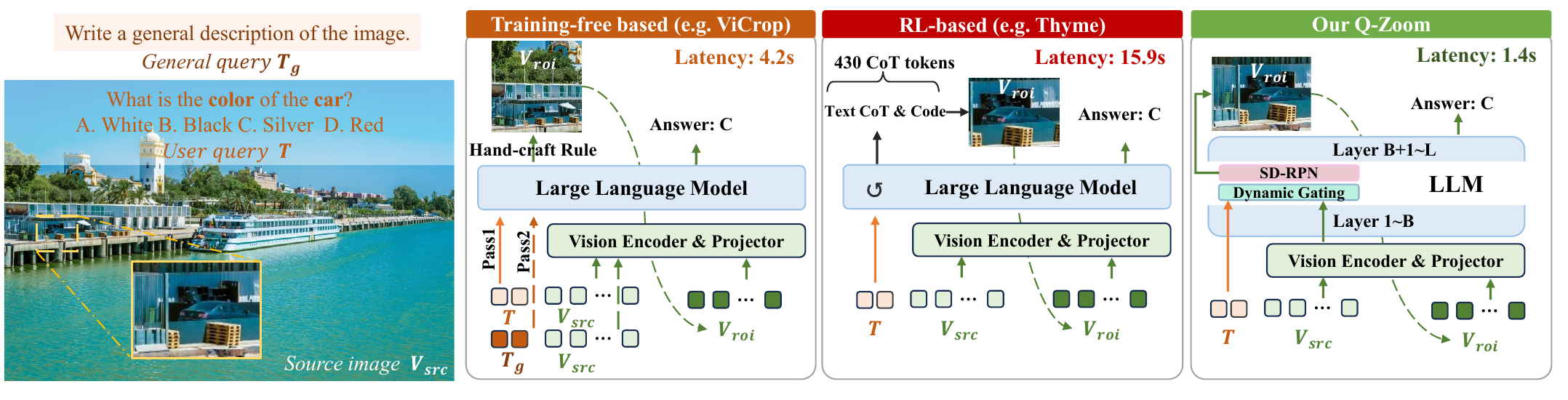}
	\caption{\textbf{Comparison of adaptive high-resolution perception paradigms.} Training-free methods rely on handcrafted contrastive rules, requiring multiple redundant prefilling passes. RL-based methods use the LLM to auto-regressively generate code or coordinates to find the RoI. Our \textbf{Q-Zoom} framework operates directly on the intermediate feature space during a single prefilling pass, yielding superior efficiency.}
	\label{fig:teaser_pipeline}
	\vspace{-4mm}
\end{figure*}

As illustrated in Figure~\ref{fig:teaser_pipeline}, recent literature attempts to tackle these perceptual inefficiencies through two primary paradigms. The first mitigates spatial sparsity via heuristic-driven, training-free pipelines~\cite{zhang2025mllms,zhong2025focus,shen2025zoomeye}. These methods exploit the MLLM's internal cross-attention to identify Regions-of-Interest (RoIs) on the fly, which are then cropped and re-encoded. However, because extracting these attention maps requires redundant prefilling passes or expensive auto-regressive decoding, they severely bottleneck inference efficiency and struggle to generalize due to rigid rules.
A second branch reformulates adaptive perception through an Reinforcement Learning (RL) based Think-with-Image  paradigm~\cite{openai2025thinking,zhang2025thyme,zheng2025deepeyes,lai2025mini}. These models identify RoIs via explicit auto-regressive reasoning or sandbox code execution. While effective at reducing visual token usage, they inadvertently shift the computational burden to the language model. Relying on lengthy Chain-of-Thought (CoT) decoding drastically extends inference latency. Furthermore, optimizing these models via RL is prohibitively expensive, data-hungry, and highly unstable.

To overcome these limitations, we propose Q-Zoom, a fully integrated, two-stage adaptive framework that addresses both query-level and spatial redundancy directly within the intermediate feature space. Inspired by findings that MLLM middle layers harbor robust visual grounding~\cite{shi2025vision, kang2025your}, we attach two lightweight sub-networks to the frozen backbone.
To eliminate query-level redundancy, the first stage introduces a Dynamic Gating Network to assess whether coarse features are sufficient. This router is optimized via a novel Consistency-Aware Sample Generation strategy, which derives deterministic routing labels by evaluating responses across a resolution trajectory, bypassing human annotations.
For queries requiring refinement, the second stage activates the Self-Distilled Region Proposal Network (SD-RPN). Operating on intermediate tokens, the SD-RPN predicts a dense heatmap to crop and re-encode only task-relevant RoIs. It is trained via a self-distillation paradigm that mines internal cross-attention maps, filters sink tokens, and applies a selective tri-state label assignment to generate pseudo-labels.
Crucially, Q-Zoom acquires these signals in a single prefilling pass. As shown in Figure~\ref{fig:teaser_pipeline}, this circumvents the redundant prefilling and sluggish auto-regressive decoding of prior methods, drastically accelerating throughput.
Finally, to resolve the spatial misalignment caused by processing the coarse global image and fine-grained local RoI, we introduce a continuous spatio-temporal positional encoding scheme. Coupled with a targeted Post-Supervised Fine-Tuning (Post-SFT) on explicitly mined hard failure cases, this teaches the LLM to seamlessly fuse local details with the global layout, restoring robust spatial reasoning.

We validate Q-Zoom across diverse base models~\cite{liu2023improvedllava, Qwen2.5-VL} on demanding Document \& OCR~\cite{mathew2020docvqa,mathew2022infographicvqa,masry2022chartqa,mishra2019ocr} and High-Resolution tasks~\cite{vstar,hrbench,zhang2024mme}. Our empirical results demonstrate that Q-Zoom establishes a new Pareto frontier in the trade-off between perceptual accuracy and computational efficiency. For example, when integrated into the Qwen2.5-VL-7B backbone, Q-Zoom not only surpasses existing training-free heuristics but also outperforms concurrent cutting-edge models. Crucially, these perceptual enhancements do not bottleneck inference latency. Through the intelligent routing of simple queries and the targeted extraction of RoIs for detail-demanding tasks, Q-Zoom exceeds the peak performance of a brute-force baseline scaled to 4,096 visual tokens, all while strictly constraining its own maximum budget to just 1,024 tokens. This elegant scaling translates to exceptional 53.0\% and 73.2\% reductions in visual token costs, alongside $2.52\times$ and $4.39\times$ accelerations in inference throughput on Document \& OCR and High-Resolution tasks, respectively.  Remarkably, Q-Zoom acts as a versatile plug-and-play module, providing orthogonal performance boosts even when integrated atop advanced RL-trained thinking models~\cite{wei2026zooming}, setting new SoTA benchmarks against concurrent cutting-edge methods.

This manuscript builds upon and extends our preliminary conference publication, SD-RPN~\cite{shi2026sdrpn}. While the original work successfully demonstrated the viability of self-distilled region proposals, it operated under a rigid pipeline without query-aware routing and suffered from spatial misalignment. To evolve this into the comprehensive \textbf{Q-Zoom} framework, our primary contributions are three-fold:
\begin{itemize}
	\item We propose \textbf{Q-Zoom}, a query-aware adaptive framework that decouples perceptual fidelity from quadratic computational costs. By introducing a lightweight \textbf{Dynamic Gating Network} alongside the SD-RPN, it eliminates both query-level and spatial redundancies in a single prefilling pass.
	\item We introduce data-efficient optimization strategies. These include a consistency-aware sample generation method to train the dynamic gate, and a self-supervised tri-state distillation paradigm for the SD-RPN, bypassing human annotations and expensive RL pipelines.
	\item We design a \textbf{continuous spatio-temporal positional encoding} scheme coupled with targeted Post-SFT to seamlessly fuse dense local RoIs with the coarse global layout. Extensive evaluations on the latest SOTA architectures demonstrate that Q-Zoom establishes a new dominant Pareto frontier in accuracy and efficiency.
\end{itemize}

	\section{Related Works}

\subsection{General Perception in MLLMs}

Classic Multimodal Large Language Model (MLLM) architectures~\cite{liu2023llava,dai2023instructblip,li2023blip} generally standardize visual inputs to a fixed resolution, employing a vision encoder aligned with the language space to produce a static number of visual tokens. For instance, methods utilizing a Q-Former~\cite{li2023blip} attempt to compress visual representations into a strict, pre-set token budget. Conversely, the LLaVA series~\cite{liu2023llava, liu2023improvedllava} adopts the dense, uncompressed token sequence directly from the vision encoder, projecting it straight into the LLM's feature space. This latter approach has gradually emerged as the mainstream paradigm in modern MLLMs due to its architectural simplicity and empirical effectiveness. 
However, standard vision encoders~\cite{radford2021learning,tschannen2025siglip,zhai2023sigmoid,bolya2025perception} (e.g., CLIP ViT~\cite{radford2021learning}) are typically pre-trained at low resolutions (e.g., $224 \times 224$ or $336 \times 336$). This training prior fundamentally bottlenecks their ability to directly encode high-resolution images, severely limiting the fine-grained perceptual capabilities of the resulting MLLMs. To tackle this limitation, recent literature has explored several distinct evolutionary pathways. 

One branch of research seeks to integrate auxiliary high-resolution vision encoders~\cite{lu2024deepseek,vasu2025fastvlm,luo2024feast,zhao2025mg,li2024mini}, such as SAM~\cite{kirillov2023segment} or ConvNeXt~\cite{liu2022convnet,radford2021learning}, to compensate for the spatial deficiencies of standalone, low-resolution ViTs. Another prominent line of work addresses the resolution gap by spatially partitioning high-resolution inputs into multiple localized patches~\cite{zhang2024llava,huang2024mini,chen2024internvl,liu2024llavanext,cha2024honeybee,li2024llava,an2025llava,wu2024deepseek}. These patches are encoded independently and subsequently concatenated before being fed to the LLM, a strategy widely popularized by the AnyRes~\cite{chen2024internvl,li2024llava} mechanism. A third alternative~\cite{Qwen2-VL,Qwen2.5-VL,lu2025ovis2} focuses on natively adapting the vision encoder to process higher, variable resolutions, producing a dynamic number of visual tokens proportional to the input size. For example, the Qwen2-VL series~\cite{Qwen2-VL} adopts a native dynamic resolution paradigm, fine-tuning the vision encoder using a NaViT-style architecture~\cite{dehghani2023patch} to seamlessly process arbitrary aspect ratios. 

\subsection{Query-aware Perception in MLLMs}

Building upon general perception frameworks, recent studies demonstrate that query-aware designs offer a more efficient and effective alternative to brute-force resolution scaling. The core principle of this paradigm is to first identify task-relevant RoIs using a coarse, low-resolution visual input, and subsequently re-encode only these cropped regions at a higher resolution. Based on their optimization strategies, existing query-aware methodologies can be broadly categorized into three paradigms: training-free, Supervised Fine-Tuning (SFT), and Reinforcement Learning (RL).

\textbf{Training-free methods}~\cite{liu2025hide,ge2025focusing,zhang2025mllms,shen2025zoomeye,liu2025seeing} rely on handcrafted heuristics to extract RoIs without updating model weights. For instance, approaches like ViCrop~\cite{zhang2025mllms} compute contrastive cross-attention maps between generic and task-specific text prompts to localize relevant visual evidence. 
However, deriving these attention signals intrinsically requires multiple redundant prefilling passes or computationally heavy auto-regressive decoding steps. 
\textbf{SFT-based methods}~\cite{jiang2025token,shi2025scaling,team2025hypervl} attempt to bypass these inference delays by teaching the MLLM to explicitly predict RoI heatmap or call external tools. This approach, however, demands the curation of massive, expensive datasets containing paired question-and-annotation coordinates. Furthermore, fully fine-tuning the LLM backbone on these specialized dense-localization datasets is computationally prohibitive and risks catastrophic forgetting, thereby degrading the foundational generalizability of the base MLLM.
\textbf{RL-based methods}~\cite{zheng2025deepeyes,zhang2025thyme,lai2025mini,liu2025faithfulness,yang2025thinking} have recently emerged to reformulate fine-grained perception into an autonomous ``Think-with-Image'' paradigm. These models are optimized via reinforcement learning to iteratively deduce visual sufficiency and locate RoIs. While effective at reducing overall visual token usage or improving the overall performance, optimizing the entire MLLM via RL incurs exorbitant GPU memory costs, suffers from training instability, and heavily depends on massive, proprietary teacher models to generate reliable reward signals. More critically, during inference, these methods inadvertently shift the computational burden from the vision encoder to the language model. They rely on lengthy Chain-of-Thought (CoT) decoding stages to ``think'' prior to answering, which dramatically inflates inference latency. Although recent latent thinking paradigms~\cite{wang2025monet,li2025latent} attempt to compress these reasoning trajectories in the hidden space, they inevitably impose a strict ceiling on the model's ultimate perceptual performance.

	\section{Method}

\subsection{Preliminaries}
\label{sec:preliminaries}

In widely adopted LLaVA-style architectures, a Multimodal Large Language Model (MLLM) typically comprises three core components: a vision encoder $\mathcal{E}_v$, a vision-language projector $\mathcal{P}$, and a Large Language Model (LLM) backbone $\mathcal{L}$ with $L$ transformer layers. Initially, the vision encoder extracts features from the raw input image $x_v$, which the projector then maps into the LLM's embedding space. We denote this initial sequence of visual embeddings as $\mathbf{H}^0_v = \mathcal{P}(\mathcal{E}_v(x_v))$, where the superscript $0$ indicates the input embedding layer. 

During the highly parallelized prefilling stage, the LLM processes these visual embeddings alongside textual tokens (e.g., system prompt $\mathbf{H}^0_{sys}$ and user query $\mathbf{H}^0_{user}$). The final layer's output for this combined context is computed as:
\begin{equation}
	\mathbf{H}^{L}_{context} = \mathcal{L}([\mathbf{H}^0_{sys}, \mathbf{H}^0_v, \mathbf{H}^0_{user}]),
\end{equation}
where $[\cdot, \cdot]$ denotes sequence concatenation. Following contextual encoding, the model generates the response via an auto-regressive decoding stage. At step $t$, the next-token probability distribution is conditioned on the preceding context:
\begin{equation}
	P(y_t | x_v, x_t, y_{<t}) = \text{Softmax}\left(\mathbf{W}_{head} \mathbf{h}^L_t\right),
\end{equation}
where $\mathbf{h}^L_t$ is the $L$-th layer's hidden state at step $t$, and $\mathbf{W}_{head}$ is the language modeling head. Due to highly parallelized matrix operations, the prefilling stage is substantially faster than auto-regressive decoding for an equivalent token count.

\subsection{Adaptive Dynamic Gating Mechanism}
\label{sec:dynamic_gating}

\begin{table}[t]
	\setlength{\tabcolsep}{2.5pt} 
	\centering
	\caption{Performance and throughput comparison of Qwen2.5-VL 7B on Document and Vision-Centric benchmarks under different maximum visual token limits. Throughput is measured in samples/second on a single RTX A6000 GPU.}
	\label{tab:pre_res_vs_acc}
	\small
	\begin{tabular}{@{}l c ccccc@{}}
		\toprule
		\begin{tabular}[c]{@{}l@{}}\textbf{Max}\\ \textbf{Limit}\end{tabular} & 
		\begin{tabular}[c]{@{}c@{}}\textbf{Through-}\\ \textbf{put}\end{tabular} & 
		\begin{tabular}[c]{@{}c@{}}\textbf{DocVQA}\\ val\end{tabular} & 
		\begin{tabular}[c]{@{}c@{}}\textbf{OCRBench}\\ test\end{tabular} & 
		\begin{tabular}[c]{@{}c@{}}\textbf{InfoVQA}\\ val\end{tabular} & 
		\textbf{V*} & 
		\textbf{Ave.} \\
		\midrule
		512 & 4.6 & 90.9 & 83.3 & 69.1 & 62.8 & 76.5 \\
		2048 & 2.2 & 94.7 & 84.8 & 80.4 & 72.3 & 83.1 \\
		\bottomrule
	\end{tabular}
	\vspace{-3mm}
\end{table}


\begin{figure*}[t]
	\centering
	\begin{subfigure}[b]{0.52\linewidth}
		\centering
		\includegraphics[width=\linewidth]{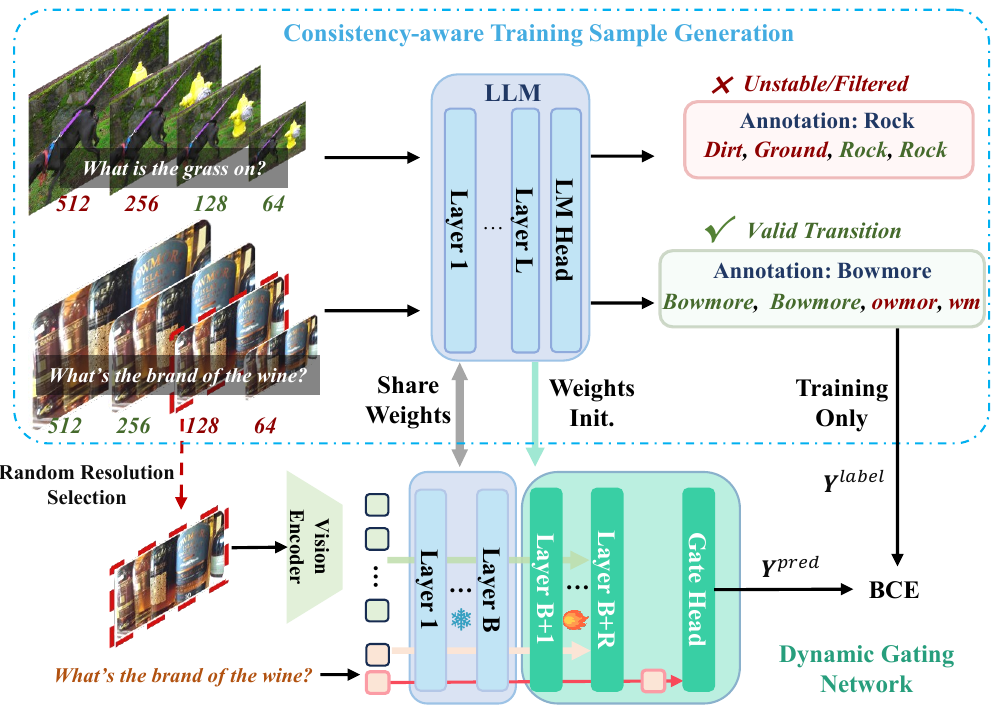}
		\caption{Consistency-aware Training \& Gating Mechanism}
		\label{fig:Dynamic_Gating_Mechanism}
	\end{subfigure}
	\hfill 
	\begin{subfigure}[b]{0.47\linewidth}
		\centering
		\includegraphics[width=\linewidth]{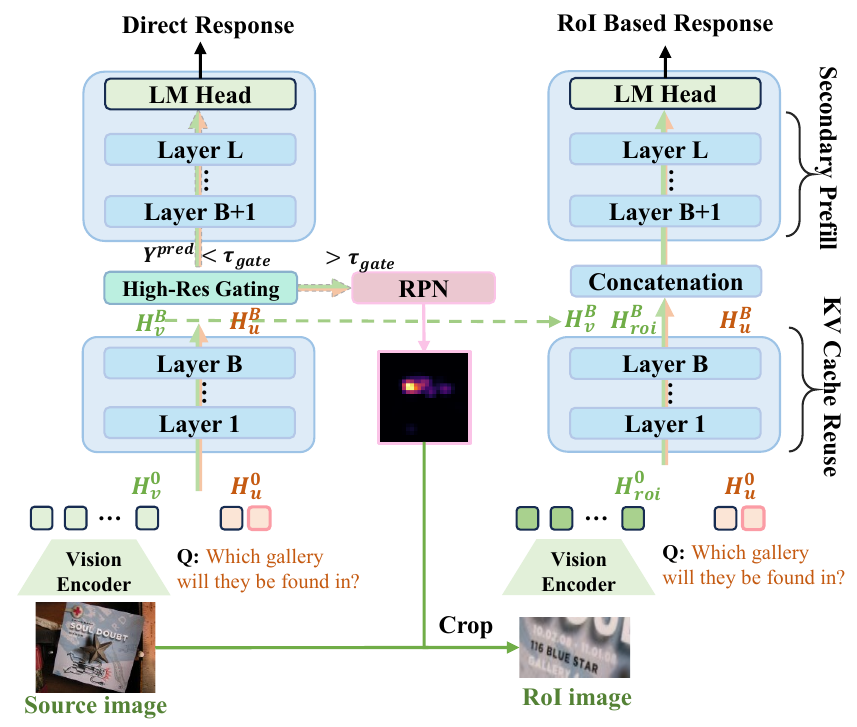}
		\caption{Adaptive Two-Stage Inference Pipeline}
		\label{fig:inference_pipeline}
	\end{subfigure}
	
	\caption{\textbf{Overview of the proposed Adaptive High-Resolution Perception Framework.} \textbf{(a)} The framework derives robust supervisory signals through consistency-aware generation to train a lightweight gating module. \textbf{(b)} During inference, the gate dynamically evaluates the textual query. It routes simpler queries for direct, accelerated generation using coarse features, while triggering the SD-RPN for complex queries to extract targeted high-resolution regions.}
	\label{fig:joint_pipeline}
	\vspace{-4mm}
\end{figure*}
Visual resolution profoundly dictates MLLMs' fine-grained perception~\cite{liu2023llava, liu2024llavanext, chen2024internvl, Qwen2-VL}, but scaling it imposes a quadratic computational bottleneck. Table~\ref{tab:pre_res_vs_acc} illustrates this trade-off for Qwen2.5-VL 7B~\cite{Qwen2.5-VL}: restricting the input from 2,048 to 512 tokens roughly doubles throughput while preserving over 90\% of relative accuracy (76.5\% vs. 83.1\%). This reveals that most queries can be resolved using coarse context, making uniform high-resolution processing highly wasteful. Therefore, we formulate high-resolution perception as a conditional routing problem: can a binary classifier dynamically predict if a specific query $(x_v, x_t)$ necessitates high-resolution refinement? As illustrated in Figure~\ref{fig:Dynamic_Gating_Mechanism}, our framework achieves this via a Consistency-aware Training Sample Generation pipeline for robust supervision, and a lightweight gating network integrated directly into the MLLM.

\subsubsection{Consistency-aware Training Sample Generation}
\label{sec:label_construction}

A naive approach assigns refinement labels based solely on the correctness of a single low-resolution response. However, MLLM performance is also influenced by intrinsic hallucinations or ambiguous queries, making such labels highly noisy. To extract clean supervisory signals, we propose a consistency-aware sample generation strategy that evaluates responses across a monotonically increasing resolution trajectory $\mathcal{R} = \{r_1, r_2, \dots, r_k\}$, yielding predictions $\{y_{r_1}, y_{r_2}, \dots, y_{r_k}\}$ (Figure~\ref{fig:Dynamic_Gating_Mechanism}, top). We apply a strict heuristic: response accuracy across resolutions should approximate a Heaviside step function. We only accept valid transition cases where the model fails at lower resolutions but succeeds at higher ones (Figure~\ref{fig:Dynamic_Gating_Mechanism}, bottom). Unstable cases where the model succeeds at low resolutions but fails at higher ones—are discarded. This consistency check guarantees that visual resolution is the deterministic factor governing correctness.


For filtered samples, we construct training pairs by randomly selecting a resolution $r \in \mathcal{R}$ to produce $x_v^{r}$. The binary gating label $Y^{label} \in \{0, 1\}$ is assigned based on the model's proficiency at $r$. If the response is incorrect, the sample supervises the Need-Refine class ($Y^{label} = 1$), triggering the RoI branch. If correct, it supervises the No-Refine class ($Y^{label} = 0$), bypassing redundant processing. This transforms multi-resolution consensus into robust binary targets, forcing the gate to learn whether additional local visual detail will tangibly change the answer's quality.

\subsubsection{Gating Network Architecture and Optimization}
\label{sec:gating_architecture}

To ensure that the routing decision remains computationally lightweight while maintaining high query awareness, we construct the dynamic gating network upon the intermediate representations of the base MLLM. Following the efficient parameter-reuse paradigm established in our preliminary work~\cite{shi2026sdrpn}, the gating module, denoted as $\mathcal{G}$, is instantiated using the pre-trained weights from layers $B+1$ to $B+R$ of the original LLM backbone. 

During the prefilling stage, the concatenated sequence of visual and textual tokens is processed through the first $B$ frozen layers of the LLM to yield the intermediate hidden states, $\mathbf{H}^B_{context}$. These representations are subsequently routed through the $R$ tunable layers of the gating network to produce the updated gating representations, $\mathbf{H}^{B+R}_{gate} = \mathcal{G}(\mathbf{H}^B_{context})$. 
To formulate a routing decision that encapsulates the task semantics, we explicitly isolate the hidden state corresponding to the final token of the user's query, denoted as $\mathbf{H}^{B+R}_{gate}[-1]$. Because the causal masking of the transformer's self-attention mechanism strictly propagates historical context forward, this terminal token inherently aggregates the full semantic intent of the question alongside the preceding visual evidence. A linear projection head, $LP_{gate}$, followed by a sigmoid activation function, $\sigma$, maps this query-aware feature to a continuous refinement probability, $Y^{pred}$. The entire gating module is then optimized via a standard Binary Cross-Entropy (BCE) loss against the deterministic binary label $Y^{label}$. The forward procedure and training objective are formally defined as:
\begin{equation}
	\begin{aligned}
		\mathbf{H}^{B+R}_{gate} &= \mathcal{G}(\mathbf{H}^B_{context}), \\
		Y^{pred} &= \sigma(LP_{gate}(\mathbf{H}^{B+R}_{gate}[-1])), \\
		\mathcal{L}_{gate} &= \text{BCE}(Y^{pred}, Y^{label}).
	\end{aligned}
\end{equation}

During inference, this architecture establishes an adaptive, input-conditioned computation pathway, as illustrated in Figure~\ref{fig:inference_pipeline}.  We introduce a predefined confidence threshold, $\tau_{gate}$, to explicitly control the trade-off between perception accuracy and inference efficiency. If the predicted probability indicates that the initial coarse-resolution input provides sufficient visual context ($Y^{pred} < \tau_{gate}$), the gating branch is bypassed. The MLLM seamlessly resumes its standard forward pass, feeding the intermediate states $\mathbf{H}^B_{context}$ through the remaining frozen layers of the backbone. Conversely, if $Y^{pred} \ge \tau_{gate}$, the gate identifies a critical insufficiency in the coarse visual evidence. This immediately suspends the standard generation pipeline and triggers the specialized RoI extraction module detailed in the following section.

\subsection{Self-Distilled Region Proposal Network}
\label{sec:sd_rpn}

\begin{figure*}[t]
	\centering
	\includegraphics[width=0.85\linewidth]{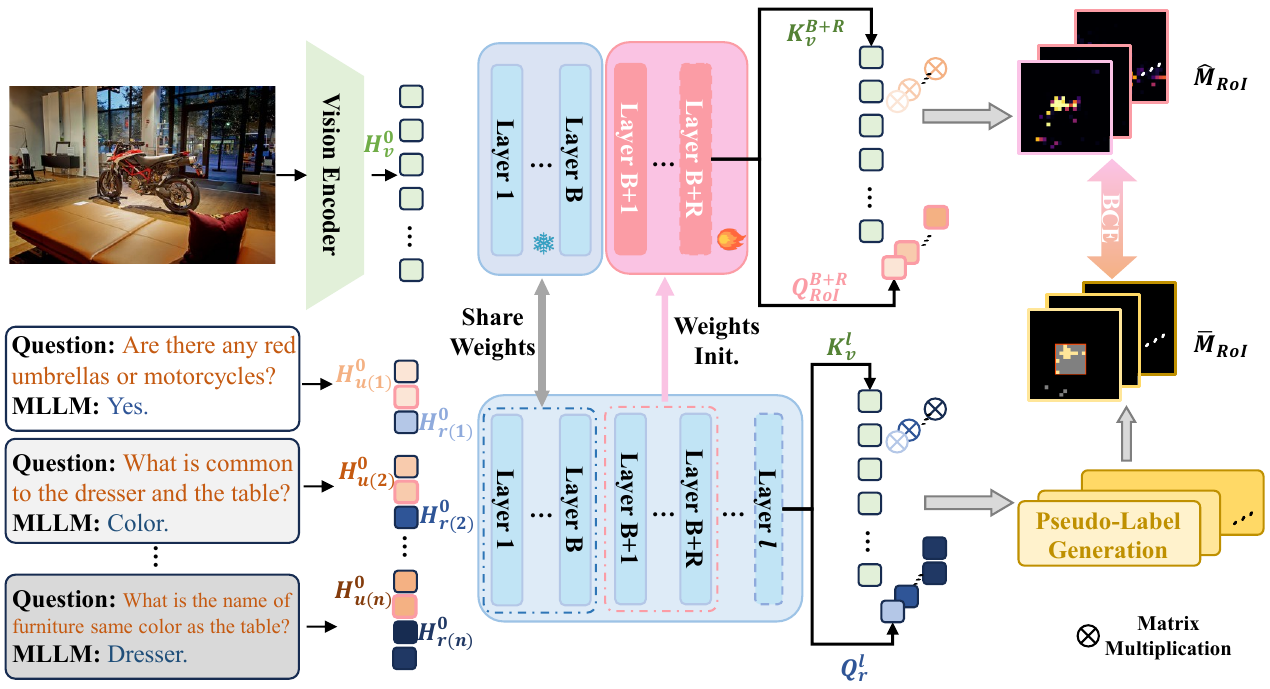}
	\caption{\textbf{Overview of the conditional Region-of-Interest extraction pipeline.} When triggered by the dynamic gating module, the SD-RPN (top) leverages shared intermediate features from the frozen backbone to efficiently generate a dense spatial heatmap. During the training phase (bottom), the network is optimized through a self-distillation paradigm, utilizing denoised cross-modal attention maps from the base MLLM as supervisory pseudo-labels. Superscripts indicate network depth (layers), while subscripts denote the modality or token origin. System prompts are excluded for visual clarity.}
	\label{fig:framework}
	\vspace{-5mm}
\end{figure*}


For complex queries that trigger the refinement pathway ($Y^{pred} \ge \tau_{gate}$), scaling the entire image incurs a severe quadratic computational bottleneck. Instead, we deploy our Self-Distilled Region Proposal Network (SD-RPN) to spatially isolate crucial visual evidence (Figure~\ref{fig:framework}). The SD-RPN dynamically localizes the Region of Interest (RoI) directly from the intermediate feature space. This localized region is then cropped from the high-resolution source image and re-encoded, providing precise, fine-grained context for final response generation.

\subsubsection{Lightweight RoI Prediction via Branched Feature Reuse}
\label{sec:roi_prediction}

Following recent findings that MLLMs' intermediate layers harbor robust visual grounding capabilities~\cite{kang2025your,shi2025vision}, we design the SD-RPN as a lightweight branch operating on the frozen backbone's intermediate features. Comprising $R$ transformer blocks initialized with pre-trained weights from layers $B+1$ to $B+R$, it structurally parallels the gating network. For notational clarity, we formalize the inference process under a single-turn conversation setting.

During the initial prefilling stage, the RPN inherits the intermediate hidden states $\mathbf{H}^B_{context}$ computed by the frozen backbone, processing them through its first $R-1$ tunable layers to yield the localized hidden states, $\mathbf{H}^{B+R-1}_{rpn}$. To predict the dense RoI map $\hat{\mathbf{M}}_{\text{RoI}}$, we repurpose the self-attention mechanism of the final ($R$-th) block into a specialized spatial prediction head. Specifically, from the sequence $\mathbf{H}^{B+R-1}_{rpn}$, we isolate the hidden state of the final user query token, denoted as $\mathbf{H}^{B+R-1}_{u}[-1] \in \mathbb{R}^{1 \times d}$, alongside the dense visual feature sequence $\mathbf{H}^{B+R-1}_v \in \mathbb{R}^{HW \times d}$, where $H$ and $W$ represent the spatial dimensions of the encoded feature map. 
Rather than introducing new, randomly initialized parameters, these elements are mapped into a shared latent space via the projection matrices ($LP_q$ and $LP_k$) native to the $R$-th attention layer in RPN. This seamlessly leverages the model's pre-aligned cross-modal semantic space:
\begin{equation}
	\begin{aligned}
		\label{eq:qk_proj}
		\mathbf{Q}_{\text{RoI}} &= LP_q(\text{Norm}(\mathbf{H}^{B+R-1}_u[-1])), \\
		\mathbf{K}_v &= LP_k(\text{Norm}(\mathbf{H}^{B+R-1}_v)),
	\end{aligned}
\end{equation}
where $\text{Norm}(\cdot)$ denotes layer normalization~\cite{ba2016layer, zhang2019root}. The spatial heatmap is derived by computing the inner product:
\begin{equation}
	\label{eq:qk_mat}
	\hat{\mathbf{M}}_{\text{RoI}} = \mathbf{Q}_{\text{RoI}} \mathbf{K}_v^\top.
\end{equation}
For mathematical brevity, the multi-head dimension is omitted; in practice, attention scores are computed independently per head and subsequently averaged.

To reliably segment foreground visual evidence, the dense map $\hat{\mathbf{M}}_{\text{RoI}}$ is activated via a sigmoid ($\sigma$), reshaped into a 2D spatial grid ($\gamma$), smoothed with a Gaussian filter ($\mathcal{G}$), and binarized using a confidence threshold ($\tau_{roi}$):
\begin{equation}
	\label{eq:fg_filtering}
	\mathcal{B}(x,y) = \begin{cases}
		1, & \text{if } \mathcal{G}(\gamma(\sigma(\hat{\mathbf{M}}_{\text{RoI}})))(x,y) > \tau_{roi}, \\
		0, & \text{otherwise},
	\end{cases}
\end{equation}
where $(x, y)$ represents the spatial coordinates. 
We compute the minimal axis-aligned bounding box $\text{bbox}(\cdot)$ enclosing the activated foreground in $\mathcal{B}$. This directs the cropping of a localized sub-image $x_{v_{\text{roi}}}$, which is re-encoded to extract fine-grained embeddings:
\begin{equation}
	b_{\text{roi}} = \text{bbox}(\mathcal{B}), \quad \mathbf{H}_{v_{\text{roi}}}^0 =  \mathcal{P}(\mathcal{E}_v(x_{v_{\text{roi}}})).
\end{equation}

To integrate this local evidence efficiently, we employ an optimized partial-prefill strategy utilizing prefix KV-cache reuse (Figure~\ref{fig:inference_pipeline}). Because the high-resolution RoI tokens ($\mathbf{H}_{v_{\text{roi}}}^0$) are inserted just before the textual query ($\mathbf{H}^0_{user}$), the prefix context (system prompt and coarse visual features) remains mathematically unchanged up to layer $B$. Thus, we directly retrieve their cached representations, $\mathbf{H}^B_{sys}$ and $\mathbf{H}^B_v$. Only the new RoI and shifted user tokens are forwarded through the first $B$ layers:
\begin{equation}
	[\mathbf{H}^B_{v_{\text{roi}}}, \mathbf{H}^B_{user}] = \mathcal{L}_{1 \to B}([\mathbf{H}_{v_{\text{roi}}}^0, \mathbf{H}^0_{user}]).
\end{equation}
These states are concatenated with the cached prefix at layer $B$ and passed through the remaining layers ($B+1$ to $L$) to generate the detail-oriented response:
\begin{equation}
	\mathbf{H}^{L}_{context} = \mathcal{L}_{B+1 \to L}([\mathbf{H}^B_{sys}, \mathbf{H}^B_v, \mathbf{H}^B_{v_{\text{roi}}}, \mathbf{H}^B_{user}]).
\end{equation}
This caching bypasses redundant re-encoding of the coarse visual context, accelerating the secondary prefilling stage.

\subsubsection{Training SD-RPN via Self-Distillation}
\label{sec:sd_rpn_training}

\begin{figure*}[t]
	\centering
	
	\begin{minipage}[c]{0.73\linewidth}
		\centering
		\includegraphics[width=\linewidth]{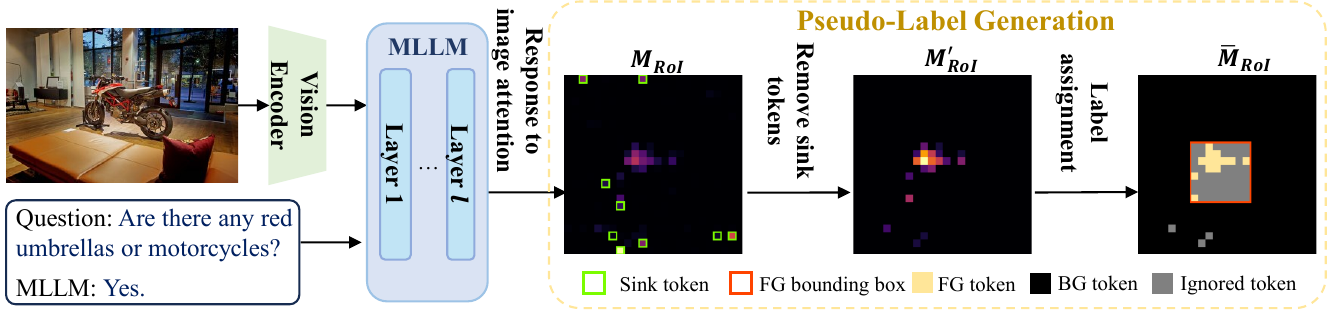}
		\vspace{-2mm} 
		\caption{\textbf{Overview of our pseudo-label generation pipeline.} Raw attention maps from the MLLM are denoised by removing sink tokens, followed by a tri-state label assignment that isolates high-confidence foreground (FG) and background (BG) tokens while ignoring ambiguous intermediate regions. Layer index is omitted for brevity.}
		\label{fig:pseudo_label_pipeline}
	\end{minipage}%
	\hfill 
	\begin{minipage}[c]{0.25\linewidth}
		\centering
		\includegraphics[width=\linewidth]{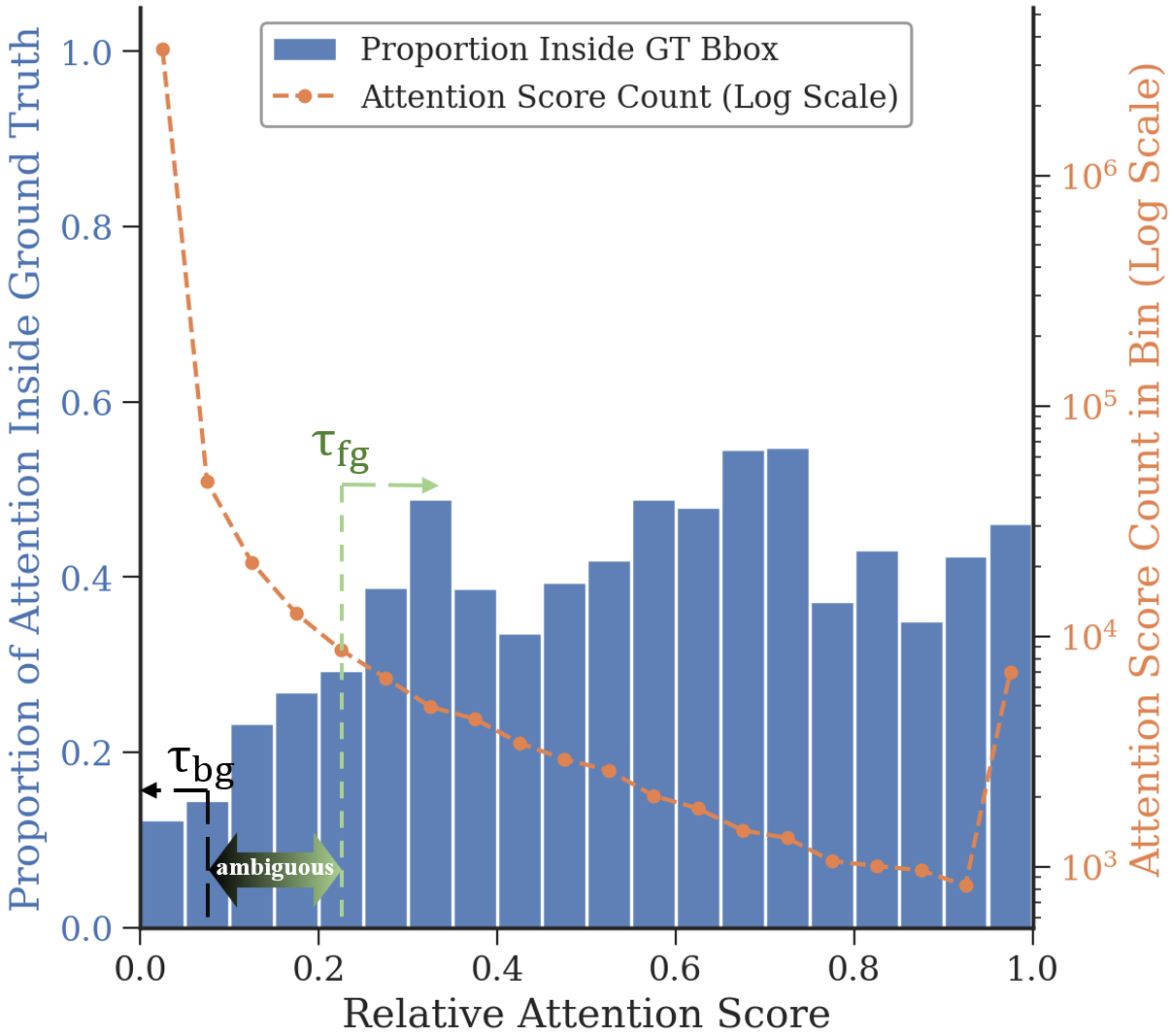}
		\vspace{-5mm} 
		\caption{\textbf{Attention magnitude vs. Localization accuracy.} }
		\label{fig:attn_loc}
	\end{minipage}
	
	\vspace{-4mm}
\end{figure*}
MLLMs' internal cross-attention mechanisms inherently possess strong visual grounding capabilities. By refining these signals, we construct high-quality pseudo-labels to supervise the SD-RPN, entirely eliminating reliance on external localization data.

\paragraph{Extracting Raw Grounding Signals}
We extract cross-modal attention weights from a designated middle layer $l$ during a standard forward pass. For a single attention head, the raw RoI map $\mathbf{M}_{\text{RoI}}^{l} \in \mathbb{R}^{H \times W}$ encapsulates each visual token's aggregated importance to the textual response:
\begin{equation}
	\mathbf{M}_{\text{RoI}}^{l} = \frac{1}{N_t} \sum_{i=1}^{N_t} \mathbf{A}_{i}^{l}, \quad \text{where} \quad \mathbf{A}^l= \text{softmax}\left(\frac{\mathbf{Q}_t^{l}(\mathbf{K}_v^{l})^\top}{\sqrt{d}}\right),
\end{equation}
where $\mathbf{Q}_t^{l} \in \mathbb{R}^{N_t \times d}$ and $\mathbf{K}_v^{l} \in \mathbb{R}^{(H \times W) \times d}$ denote the query and key embeddings of the response and visual tokens.

\paragraph{Robust Pseudo-Label Construction}
Directly utilizing $\mathbf{M}_{\text{RoI}}^{l}$ as a dense supervisory signal is suboptimal because raw attention distributions are notoriously noisy. As illustrated in Fig.~\ref{fig:pseudo_label_pipeline}, they frequently suffer from high-activation artifacts in background regions and fragmented activation across the foreground object. We therefore introduce a robust pseudo-label generation pipeline to systematically denoise this signal.

The first source of noise arises from \textit{sink tokens}---visual tokens that accumulate disproportionate attention mass despite lacking semantic relevance to the grounded object. Following observations in recent studies~\cite{darcet2023vision, kang2025see}, these tokens consistently exhibit anomalously large $\text{L}_2$-norms in their feature representations. We filter them by applying a predefined norm threshold $\tau_{\text{norm}}$, yielding a denoised attention map $\mathbf{M}'_{\text{RoI}}$:
\begin{equation}
	(\mathbf{M}'_{\text{RoI}})_j = \begin{cases} 
		0, & \text{if } \|(\mathbf{H}_{v})_j\|_2 > \tau_{\text{norm}}, \\ 
		(\mathbf{M}_{\text{RoI}})_j, & \text{otherwise}. 
	\end{cases}
\end{equation}

Second, we resolve the ambiguity of foreground-background margins. Empirical analysis detailed in Fig.~\ref{fig:attn_loc} on the TextVQA dataset reveals that while tokens with extreme relative attention scores ($a_j/a_{\max}$) reliably correlate with ground-truth foreground or background, numerous tokens fall into a highly ambiguous middle range. 
To mitigate this, we formulate a selective tri-state classification strategy. 
We define a high-confidence foreground set $\mathcal{S}_{fg} = \{ j \mid a_j \ge \tau_{fg}\, a_{\max}\}$ and establish a minimal bounding box $\mathcal{B}_{fg}$ that spatially encloses these foreground tokens. To prevent incomplete object activation from incorrectly penalizing the network, any token residing inside $\mathcal{B}_{fg}$ that is not explicitly in $\mathcal{S}_{fg}$ is assigned an ignore label. The background set $\mathcal{S}_{bg}$ is strictly constrained to tokens outside $\mathcal{B}_{fg}$ with low attention scores ($a_j \le \tau_{bg}\, a_{\max}$). The final discrete pseudo-label map, $\bar{\mathbf{M}}_{\text{RoI}}$, is thus constructed as:
\begin{equation}
	\label{eq:pseudo_label_making}
	(\bar{\mathbf{M}}_{\text{RoI}})_j = \begin{cases} 
		1, & \text{if token } j \in \mathcal{S}_{fg}, \\ 
		0, & \text{if token } j \in \mathcal{S}_{bg}, \\ 
		-1, & \text{otherwise (ignored).} 
	\end{cases}
\end{equation}


To facilitate multi-turn interactions while bypassing computationally expensive decoding steps during training, we extract the hidden states $\mathbf{H}^l$ from SD-RPN's penultimate layer $(l=B+R-1)$  across an $n$-turn dialogue:
\begin{equation}
	\mathbf{H}^l =[\mathbf{H}^l_{\text{sys}}, \mathbf{H}^l_v, \mathbf{H}^l_{u(1)}, \mathbf{H}^l_{r(1)}, \ldots, \mathbf{H}^l_{u(n)}, \mathbf{H}^l_{r(n)}].
\end{equation}
We isolate each user query's terminal token and concatenate them into an aggregated query tensor $\mathbf{H}^l_{\text{RoI}}$:
\begin{equation}
	\mathbf{H}^l_{\text{RoI}} = \text{concat}(\mathbf{H}^l_{u(1)}[-1], \ldots, \mathbf{H}^l_{u(n)}[-1]).
\end{equation}
These queries and dense visual states $\mathbf{H}_v$ are projected (Eq.~\ref{eq:qk_proj}) to compute the multi-turn RoI map $\hat{\mathbf{M}}_{\text{RoI}}$ (Eq.~\ref{eq:qk_mat}). The RPN is optimized via a selective BCE loss $\mathcal{L}_{\text{RPN}}=\text{BCE}(\hat{\mathbf{M}}_{\text{RoI}}, \bar{\mathbf{M}}_{\text{RoI}})$, computing gradients over valid tokens.

\subsection{Spatio-Temporal Alignment and Targeted Fine-Tuning}
\label{sec:post_sft}

\begin{figure}[t]
	\centering
	\includegraphics[width=0.75\linewidth]{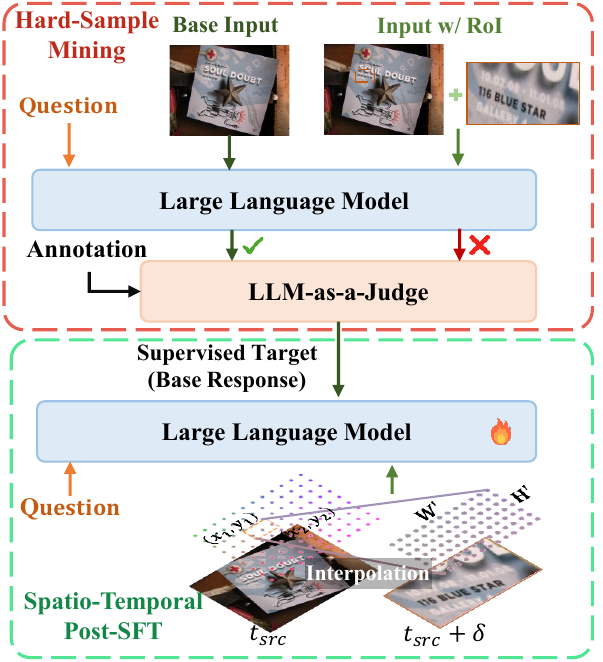}
	\caption{\textbf{Overview of the Spatio-Temporal Alignment and Targeted Post-SFT pipeline.} The vision encoder and projector are omitted for visual brevity.}
	\label{fig:post_sft_pipeline}
	\vspace{-4mm}
\end{figure}

While the extraction of high-resolution RoIs effectively isolates fine-grained visual details, it detaches the cropped region from its broader spatial context. For MLLMs equipped with Multimodal Rotary Positional Embeddings (MRoPE)~\cite{Qwen2-VL, Qwen2.5-VL}, processing the coarse source image and the localized RoI as two independent visual sequences often induces spatial misalignment. Consequently, the model struggles to map the RoI back to its original physical location, leading to degraded performance on tasks requiring global spatial reasoning (e.g., determining relative object placement). To resolve this, we propose a continuous spatio-temporal positional encoding scheme coupled with a targeted post-supervised fine-tuning (Post-SFT) strategy, as illustrated in Figure~\ref{fig:post_sft_pipeline}.

\textbf{Continuous Spatio-Temporal Alignment.} To reintegrate the local RoI into the global spatial layout, we explicitly inject the source coordinates into the RoI tokens through a dual-axis positional adjustment: \textit{Temporal Shift} and \textit{Spatial Interpolation}. 
First, as depicted on the right side of Figure~\ref{fig:post_sft_pipeline}, to logically distinguish the dense RoI tokens from the coarse source tokens sharing the same spatial footprint, thereby preventing positional collision, we assign the RoI tokens an offset temporal index $t_{\text{roi}} = t_{\text{src}} + \delta$. This operation effectively projects the high-resolution RoI onto an auxiliary temporal layer directly overlaid on the source image. Following standard MRoPE implementations, the offset $\delta$ is set to $\min(H,W)$, where $H$ and $W$ denote the spatial dimensions of the source visual feature map.
Second, to preserve semantic localization, the spatial position IDs for the RoI are derived directly from the source image's bounding box coordinates. Because the cropped RoI yields a denser grid of visual tokens ($H' \times W'$) than the equivalent region in the source image, we interpolate the sparse source coordinates to populate the dense RoI grid. 
Formally, let $\text{Embed}(t, h, w)$ denote the MRoPE function, and let $b =[x_1, y_1, x_2, y_2]$ represent the precise bounding box of the RoI normalized to the source coordinate space. The continuous spatio-temporal position embedding for an RoI token at grid index $(i, j)$ is computed as:
\begin{equation}
	\label{eq:spatio_temporal_pe}
	\begin{aligned}
		\mathbf{p}_{\text{roi}}^{(i,j)} = \text{Embed}\bigg( & t_{\text{src}} + \delta, \ y_1 + \frac{i}{H'-1}(y_2 - y_1), \\
		& x_1 + \frac{j}{W'-1}(x_2 - x_1) \bigg),
	\end{aligned}
\end{equation}
where $i \in \{0, \dots, H'-1\}$ and $j \in \{0, \dots, W'-1\}$. 
This formulation guarantees that the dense RoI tokens remain explicitly grounded within their original global coordinates.

\textbf{Targeted Post-Supervised Fine-Tuning (Post-SFT).} 
Even with rigorous positional alignment, pre-trained LLM backbones lack the inherent capacity to fuse these dual-stream (coarse global + dense local) inputs. The sudden influx of concentrated local features can distract the model, overshadowing the global context. 
To correct this contextual imbalance without fine-tuning the model on generic multimodal datasets which is computationally expensive and risks catastrophic forgetting, we construct a targeted dataset via contrastive hard-sample mining. As shown on the left side of Figure~\ref{fig:post_sft_pipeline}, we employ an LLM-as-a-Judge to evaluate parallel responses from two configurations: the original \textit{Base Model} and our un-finetuned \textit{RoI Based Model} (provided the source image plus the aligned RoI). By isolating the subset of hard samples where the Base Model answers correctly but the RoI Model fails, we capture instances of spatial misalignment and contextual distraction. 
During the Post-SFT phase, the Vision Encoder and Projector remain frozen. Only the LLM backbone is updated using this mined dataset of hard samples. 
This targeted optimization teaches the LLM how to dynamically balance and integrate high-resolution RoI features with the coarse global context, restoring robust global spatial reasoning.
	\section{Experiments}
\label{sec:experiments}

\subsection{Experiment Settings}
\label{sec:experiment_settings}

\begin{table*}[t]
	\setlength{\tabcolsep}{3pt}
	\centering
	\caption{Performance on Document \& OCR benchmarks. Dataset subscripts denote the evaluation split. Performance subscripts show the absolute improvement ($\uparrow$) over the baseline. Throughput is relative to the baseline, measured on a single NVIDIA A6000 GPU and our results are evaluated under a constraint of 576 maximum visual tokens.}
	\label{tab:main_doc_ocr}
	\begin{tabular}{@{}l c ccccc c@{}}
		\toprule
		\textbf{Methods} & \textbf{Throughput} & \textbf{DocVQA}\textsubscript{val} & \textbf{ChartQA}\textsubscript{test} & \textbf{OCRBench}\textsubscript{test} & \textbf{InfoVQA}\textsubscript{val} & \textbf{TextVQA}\textsubscript{val} & \textbf{Ave.} \\
		\midrule
		\textbf{LLaVA-1.5-7B} & \textbf{1.0$\times$} & 21.5 & 18.1 & 31.4 & 20.4 & 46.1 & 27.5 \\
		+S$^2$\venuecite{ECCV2024}{shi2024we} & 0.70$\times$ & 27.1 & 18.9 & 32.6 & \textbf{22.5} & 52.6 & 30.7 \\
		+ViCrop\venuecite{ICLR2025}{zhang2025mllms} & 0.42$\times$ & 27.0 & 20.0 & 33.2 & 21.4 & 57.2 & 31.8 \\
		+SD-RPN\venuecite{ICLR2026}{shi2026sdrpn} & 0.62$\times$ & \textbf{34.2} & 20.6 & 37.3 & 22.3 & \textbf{58.8} & 34.6 \\
		\rowcolor{headergray}
		+Q-Zoom & 0.64$\times$ & 34.0 & \textbf{20.8} & \textbf{38.0} & 22.4 & 58.4 & \textbf{34.7}\impgain{7.2} \\
		\midrule
		\textbf{LLaVA-1.5-13B} & \textbf{1.0$\times$} & 23.5 & 18.1 & 33.7 & 23.4 & 48.7 & 29.5 \\
		+S$^2$\venuecite{ECCV2024}{shi2024we} & 0.71$\times$ & 30.7 & 20.3 & 36.4 & 24.7 & 54.5 & 33.3 \\
		+ViCrop\venuecite{ICLR2025}{zhang2025mllms} & 0.39$\times$ & 30.2 & 20.1 & 36.1 & \textbf{25.9} & 60.3 & 34.5 \\
		+SD-RPN\venuecite{ICLR2026}{shi2026sdrpn} & 0.64$\times$ & \textbf{39.4} & 21.2 & \textbf{39.6} & 24.8 & \textbf{63.4} & \textbf{37.7} \\
		\rowcolor{headergray}
		+Q-Zoom & 0.66$\times$ & \textbf{39.4} & \textbf{21.3} & 39.4 & 25.1 & 62.9 & 37.6\impgain{8.1} \\
		\midrule
		\textbf{Qwen2.5-VL-3B} & \textbf{1.0$\times$} & 87.1 & 82.3 & 75.9 & 62.4 & 76.7 & 76.9 \\
		+SD-RPN\venuecite{ICLR2026}{shi2026sdrpn} & 0.49$\times$ & 89.5 & 83.8 & 76.3 & 67.0 & 79.7 & 79.3 \\
		\rowcolor{headergray}
		+Q-Zoom & 0.73$\times$ & \textbf{91.8} & \textbf{84.1} & \textbf{79.8} & \textbf{74.5} & \textbf{80.0} & \textbf{82.0}\impgain{5.1} \\
		\midrule
		\textbf{Qwen2.5-VL-7B} & \textbf{1.0$\times$} & 92.0 & 83.0 & 82.8 & 70.1 & 81.1 & 81.8 \\
		+VisionThink\venuecite{NeurIPS2025}{yang2025visionthink} & -- & 93.7 &  73.9 & 80.8 & -- & -- & -- \\
		+AdaptVision\venuecite{CVPR2026}{lin2025adaptvision} & 0.06$\times$ & 92.6 &  75.9 & 76.9 & -- & -- & -- \\
		+SD-RPN\venuecite{ICLR2026}{shi2026sdrpn} & 0.50$\times$ & 93.6 & 85.5 & 82.9 & 76.9 & \textbf{83.5} & 84.5 \\
		\rowcolor{headergray}
		+Q-Zoom & 0.81$\times$ & \textbf{94.3} & \textbf{85.6} & \textbf{85.4} & \textbf{79.4} & \textbf{83.5} & \textbf{85.6}\impgain{3.8} \\
		\midrule
		\textbf{Qwen3-VL-4B} & \textbf{1.0$\times$} & 91.3 & 84.0 & 83.1 & 68.0 & 79.2 & 81.1 \\
		+SD-RPN\venuecite{ICLR2026}{shi2026sdrpn} & 0.63$\times$ & 92.8 & 84.0 & 84.2 & 75.4 & 78.2 & 82.9 \\
		\rowcolor{headergray}
		+Q-Zoom & 0.82$\times$ & \textbf{93.4} & \textbf{85.0} & \textbf{84.6} & \textbf{77.1} & \textbf{81.4} & \textbf{84.3}\impgain{3.2} \\
		\bottomrule
	\end{tabular}
	\vspace{-4mm}
\end{table*}

\begin{table*}[t]
	\centering
	\caption{\textbf{Performance on Vision-Centric and High-Resolution benchmarks.} Dataset subscripts denote the specific evaluation split. Performance subscripts indicate the absolute improvement ($\uparrow$) of our latest version over the baseline. \textbf{Tp} denotes the relative inference throughput. Averages are computed exclusively across the four Overall metrics. Unless otherwise noted, our results are evaluated under a constraint of 4,096 maximum visual tokens. The $\dagger$ symbol denotes results directly cited from the corresponding original publications.}
	\label{tab:main_hr}
	\setlength{\tabcolsep}{3.5pt}
	\begin{tabular}{@{}l c ccc c ccc  ccc c@{}}
		\toprule
		\multirow{2}{*}{\textbf{Methods}} & \multirow{2}{*}{\textbf{Tp}} & \multicolumn{3}{c}{\textbf{V* Bench}} & \textbf{MME-RW} & \multicolumn{3}{c}{\textbf{HR-Bench 4K}} & \multicolumn{3}{c}{\textbf{HR-Bench 8K}} & \multirow{2}{*}{\textbf{Ave.}} \\
		\cmidrule(lr){3-5} \cmidrule(lr){6-6} \cmidrule(lr){7-9} \cmidrule(lr){10-12}
		& & \textbf{Attr} & \textbf{Spatial} & \textbf{Overall} & \textbf{Lite} & \textbf{FSP} & \textbf{FCP} & \textbf{Overall} & \textbf{FSP} & \textbf{FCP} & \textbf{Overall} & \\
		\midrule
		\textbf{LLaVA-1.5-7B} & 1.00$\times$ & 48.7 & 52.6 & 50.3 & \textbf{27.7} & 39.5 & 35.5 & 37.5 & 32.5 & 33.8 & 33.8 & 37.3 \\
		+S$^2$\venuecite{ECCV2024}{shi2024we} & 0.63$\times$ & 53.0 & 59.1 & 55.5 & \textbf{28.5} & 49.8 & \textbf{38.3} & 44.0 & 40.5 & \textbf{36.5} & 38.5 & 41.6 \\
		+ViCrop\venuecite{ICLR2025}{zhang2025mllms} & 0.15$\times$ & 53.9 & 50.0 & 52.4 & 27.6 & \textbf{60.8} & 34.8 & \textbf{47.8} & 38.5 & 33.8 & 36.1 & 41.0 \\
		+SD-RPN\venuecite{ICLR2026}{shi2026sdrpn} & 0.57$\times$ & \textbf{70.4} & \textbf{71.1} & \textbf{70.7} & 27.7 & 59.0 & 35.5 & 47.3 & \textbf{48.8} & 34.5 & \textbf{41.6} & \textbf{46.8} \\
		\rowcolor{headergray}
		+Q-Zoom & 0.58$\times$ & \textbf{70.4} & \textbf{71.1} & \textbf{70.7}\impgain{20.4} & 27.7\impgain{0.0} & 58.8 & 35.8 & 47.1\impgain{9.6} & 48.5 & 34.5 & 41.4\impgain{7.6} & 46.7\impgain{9.4} \\
		
		\midrule
		\textbf{LLaVA-1.5-13B} & 1.00$\times$ & 47.0 & 56.6 & 50.8 & 27.8 & 41.5 & 44.5 & 43.0 & 36.6 & 38.5 & 36.6 & 39.6 \\
		+S$^2$\venuecite{ECCV2024}{shi2024we} & 0.73$\times$ & 43.5 & 59.2 & 49.7 & \textbf{36.3} & 51.8 & 46.5 & 49.1 & 41.3 & \textbf{44.5} & 42.9 & 44.5 \\
		+ViCrop\venuecite{ICLR2025}{zhang2025mllms} & 0.21$\times$ & 47.8 & 57.9 & 51.8 & 30.1 & \textbf{66.3} & 40.3 & \textbf{53.3} & 44.5 & 37.0 & 40.6 & 44.0 \\
		+SD-RPN\venuecite{ICLR2026}{shi2026sdrpn} & 0.57$\times$ & 60.9 & \textbf{65.8} & \textbf{62.8} & 31.4 & 58.8 & \textbf{47.0} & \textbf{52.9} & \textbf{51.8} & 39.8 & \textbf{45.8} & \textbf{48.2} \\
		\rowcolor{headergray}
		+Q-Zoom & 0.58$\times$ & \textbf{61.0} & 64.5 & 61.8\impgain{11.0} & 31.3\impgain{3.5} & 59.3 & 44.8 & 52.0\impgain{9.0} & 36.8 & \textbf{51.0} & 43.9\impgain{7.3} & 47.3\impgain{7.7} \\
		
		\midrule
		\textbf{Qwen2.5-VL-3B} & 1.00$\times$ & 81.7 & 60.5 & 73.3 & 41.6 & 80.5 & 52.0 & 66.3 & 70.0 & 47.8 & 58.9 & 60.0 \\
		+SD-RPN\venuecite{ICLR2026}{shi2026sdrpn} & 0.66$\times$ & \textbf{91.3} & 65.8 & \textbf{81.2} & 41.9 & 88.0 & \textbf{58.3} & \textbf{73.1} & 79.5 & 51.0 & 65.3 & 65.4 \\
		\rowcolor{headergray}
		+Q-Zoom & 0.67$\times$ & 87.0 & \textbf{69.7} & 80.1\impgain{6.8} & \textbf{44.0}\impgain{2.4} & \textbf{88.8} & 54.8 & 71.8\impgain{5.5} & \textbf{87.3} & \textbf{55.5} & \textbf{71.4}\impgain{12.5} & \textbf{66.8}\impgain{6.8} \\
		\midrule
		\textbf{Qwen2.5-VL-7B} & 1.00$\times$ & 80.0 & 75.0 & 78.0 & 42.7 & 81.8 & 62.8 & 72.5 & 72.8 & 54.5 & 63.6 & 64.2 \\
		+Thyme\venuecite{ICLR2026}{zhang2025thyme} & 0.21$\times$ & 83.5 & \textbf{80.3} & 82.2 & \textbf{53.7} & 91.0 & 63.0 & 77.0 & 86.5 & 57.5 & 72.0 & 71.2 \\
		+DeepEyes$^\dagger$\venuecite{ICLR2026}{zheng2025deepeyes} & - & - & - & 85.6 & 50.9 & 91.3 & 59.0 & 75.1 & 86.6 & 58.5 & 72.6 & 71.1 \\
		+DeepEyesv2$^\dagger$\venuecite{ICLR2026}{hong2025deepeyesv2} & - & - & - & 81.8 & - & - & - & 77.9 & - & - & 73.8 & - \\
		+SD-RPN\venuecite{ICLR2026}{shi2026sdrpn} & 0.77$\times$ & \textbf{94.8} & 77.6 & \textbf{88.0} & 46.4 & \textbf{92.5} & 64.5 & 78.5 & 86.5 & 60.5 & 73.5 & 71.6 \\
		\rowcolor{headergray}
		+Q-Zoom & 0.86$\times$ & 89.6 & 79.0 & 85.3\impgain{7.3} & 48.0\impgain{5.3} & 91.3 & \textbf{65.8} & \textbf{78.5}\impgain{6.0} & \textbf{90.8} & \textbf{63.8} & \textbf{77.3}\impgain{13.7} & \textbf{72.3}\impgain{8.1} \\
		\midrule
		\textbf{Qwen3-VL-4B} & 1.00$\times$ & 86.1 & 76.3 & 83.2 & 47.4 & 85.8 & 63.8 & 74.8 & 73.5 & 64.0 & 68.8 & 68.6 \\
		+SD-RPN\venuecite{ICLR2026}{shi2026sdrpn} & 0.58$\times$ & \textbf{96.5} & 82.9 & 91.1 & 51.8 & \textbf{92.3} & 69.5 & \textbf{80.9} & \textbf{87.5} & \textbf{64.8} & \textbf{76.1} & 75.0 \\
		\rowcolor{headergray}
		+Q-Zoom & 0.73$\times$ & 95.7 & \textbf{85.5} & \textbf{91.6}\impgain{8.4} & \textbf{53.5}\impgain{6.1} & 89.8 & \textbf{70.8} & 80.3\impgain{5.5} & \textbf{87.5} & \textbf{64.8} & \textbf{76.1}\impgain{7.3} &
		\textbf{75.4}\impgain{6.8} \\
		\midrule
		\textbf{ZwZ-Qwen2.5-VL-7B}\venuecite{Arxiv2026}{wei2026zooming} & 1.00$\times$ & 87.0 & 82.9 & 85.3 & 52.5 & 84.0 & 64.0 & 74.0 & 73.8 & 57.0 & 65.4 & 69.3 \\
		\rowcolor{headergray}
		+Q-Zoom & 0.76$\times$ & \textbf{94.8} & \textbf{86.8} & \textbf{91.6}\impgain{6.3} & \textbf{53.2}\impgain{0.7} & \textbf{93.8} & \textbf{65.3} & \textbf{79.5}\impgain{5.5} & \textbf{92.5} & \textbf{66.3} & \textbf{79.4}\impgain{14.0} &
		\textbf{75.9}\impgain{6.6} \\
		\textbf{ZwZ-Qwen3-VL-4B}\venuecite{Arxiv2026}{wei2026zooming} & 1.00$\times$ & 89.6 & 86.8 & 88.5 & 54.3 & 85.8 & 66.3 & 76.0 & 77.0 & 65.8 & 71.4 & 72.5 \\
		\rowcolor{headergray}
		+Q-Zoom & 0.66$\times$ & \textbf{96.5} & \textbf{90.8} & \textbf{94.2}\impgain{5.7} & \textbf{54.9}\impgain{0.6} & \textbf{93.0} & \textbf{69.5} & \textbf{81.3}\impgain{5.3} & \textbf{91.3} & \textbf{69.8} & \textbf{80.5}\impgain{9.1} &
		\textbf{77.7}\impgain{5.2} \\
		\bottomrule
	\end{tabular}
	\vspace{-4mm}
\end{table*}

\paragraph{Benchmarks.} 
We evaluate our framework across two benchmark categories demanding fine-grained perception: 1) \textit{Document \& OCR} (DocVQA~\cite{mathew2020docvqa}, InfoVQA~\cite{mathew2022infographicvqa}, ChartQA~\cite{masry2022chartqa}, OCRBench~\cite{liu2024ocrbench}, and TextVQA~\cite{singh2019towards}); and 2) \textit{High-Resolution \& Vision-Centric} (V*~\cite{vstar}, MME-RealWorld~\cite{zhang2024mme}, and HR-Bench~\cite{hrbench}). For core component ablation studies, we also include General QA benchmarks~\cite{fu2023mme, chen2024we} to verify the preservation of multimodal generalizability.

\paragraph{Implementation Details.} 
Our training pipeline comprises two paradigms. First, during \textit{efficient partial tuning}, we freeze the base MLLM and optimize only the newly introduced branch parameters. The dynamic gating network and SD-RPN are trained on filtered subsets of standard VQA and document datasets using our proposed label generation strategies. Notably, we exclude extreme-resolution samples for LLaVA-series models during SD-RPN training, as their limited base resolution causes instability in pseudo-label extraction. 
Second, for \textit{targeted Post-SFT}, we fine-tune only the LLM backbone on a highly curated set of $\sim$7K hard samples. These are mined via an LLM-as-a-Judge by isolating instances where the base model succeeds but unaligned RoI integration induces failure. Because LLaVA lacks Multimodal Rotary Positional Embeddings (MRoPE), this Post-SFT stage is exclusively applied to Qwen variants. Comprehensive dataset compositions and hyperparameters are provided in the Appendix.

\paragraph{Inference Configurations.} 
For standard benchmarks, we cap Qwen's maximum visual tokens at 576 to align with LLaVA baselines and our previous implementations. To preserve dynamic aspect-ratio encoding, we relax the minimum token count (e.g., to 128), as forcing strict equality between minimum and maximum limits harms baseline performance. For resource-intensive high-resolution benchmarks, we elevate the baseline limit to 4,096 tokens to ensure fair and rigorous comparisons against competing SoTA algorithms~\cite{zheng2025deepeyes, zhang2025thyme}.

\subsection{Main Results}
We present the overall performance of our proposed framework and SoTA competitors in Table~\ref{tab:main_doc_ocr} and Table~\ref{tab:main_hr}, which detail the results on Document and OCR benchmarks, and High-Resolution and Vision-Centric benchmarks, respectively. In addition to accuracy, we report the inference throughput as a key performance metric to reflect practical computational efficiency. All experimental evaluations for our methods are conducted using the LMMS-Eval framework~\cite{zhang2024lmmsevalrealitycheckevaluation} on a single NVIDIA RTX A6000 GPU.

\textbf{Quantitative Results.} 
We evaluate Q-Zoom against paradigms including direct resolution scaling (S$^2$~\cite{shi2024we}), training-free heuristics (ViCrop~\cite{zhang2025mllms}), RL-based dynamic routing (AdaptVision~\cite{lin2025adaptvision}), and our preliminary framework (SD-RPN~\cite{shi2026sdrpn}). Across both LLaVA~\cite{liu2023llava} and Qwen architectures, Q-Zoom establishes a lead in the majority of evaluations. On LLaVA-1.5-7B, it yields a 7.2\% average gain while operating $\sim$1.5$\times$ faster than ViCrop. This efficiency gap is starkest against AdaptVision on Qwen2.5-VL-7B, where Q-Zoom achieves a staggering $>$10$\times$ speedup. This exposes a structural limitation of RL-based paradigms: their reliance on auto-regressive Chain-of-Thought decoding prior to RoI extraction severely bottlenecks throughput. (Note: AdaptVision was evaluated without vLLM~\cite{kwon2023efficient} under a 2,048 visual token limit for fair alignment with our implementation).

Compared to the preliminary SD-RPN, improvements vary by architecture. On LLaVA, gains are marginal because its low base resolution ($336\times336$) forces the gating network to almost universally trigger the RoI branch for high-resolution test images, limiting throughput improvements. Conversely, on Qwen baselines, efficiency improves by over 30\%, alongside absolute accuracy gains ranging from +3.2\% (Qwen3-VL-4B) to +5.1\% (Qwen2.5-VL-3B), directly validating our newly introduced dynamic gating and targeted fine-tuning.

Beyond document understanding, we evaluate Q-Zoom on visually intensive environments requiring small-subject detection and spatial reasoning. Standard global down-sampling strategies inherently compress and distort these fine-grained details, leading to suboptimal result. As shown in Table~\ref{tab:main_hr}, Q-Zoom outperforms SoTA methods like Thyme~\cite{zhang2025thyme} and DeepEyes~\cite{zheng2025deepeyes,hong2025deepeyesv2}. On Qwen2.5-VL-7B, our method sets a new SoTA (72.3\%), beating DeepEyes by 1.2\% and Thyme by 1.1\%. These improvements transfer robustly to Qwen3-VL architectures. Importantly, Q-Zoom is complementary to advanced reasoning paradigms. Integrated into RL-trained models (ZwZ-Qwen2.5-VL and ZwZ-Qwen3-VL~\cite{wei2026zooming}), it yields further absolute gains of 6.6\% and 5.2\%, respectively. Crucially, Q-Zoom elegantly bypasses the text-decoding bottleneck plaguing RL models like Thyme; operating at $0.86\times$ relative throughput, it runs over $4\times$ faster than Thyme while delivering superior accuracy.

\begin{figure*}[t]
	\centering
	\includegraphics[width=1.0\linewidth]{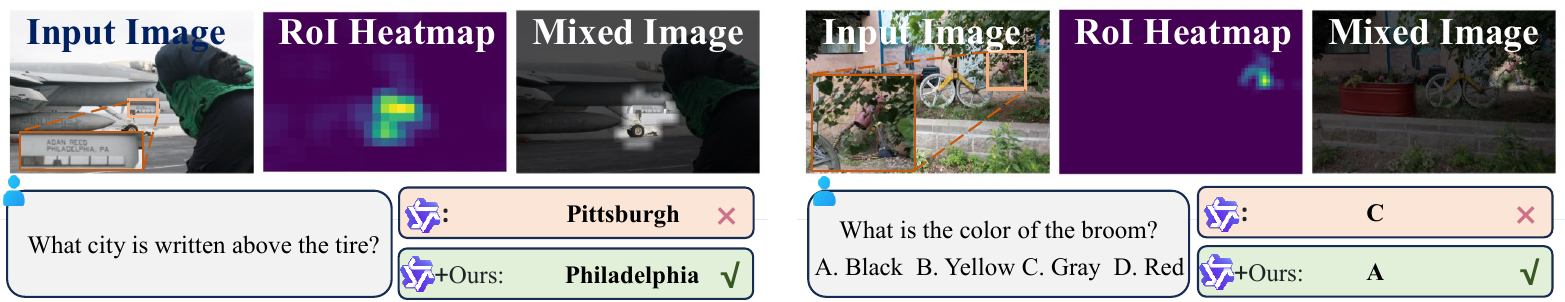} 
	\caption{
		\textbf{Qualitative comparisons on challenging examples from TextVQA (left) and V* Bench (right).} These examples highlight visually demanding scenarios where the target evidence is small or obscured. The baseline Qwen2.5-VL-7B suffers from resolution compression, whereas our Q-Zoom framework successfully leverages the SD-RPN to predict highly accurate RoI heatmaps, cropping the necessary fine-grained details to generate the correct answers.
	}
	\label{fig:main_vis}
	\vspace{-4mm}
\end{figure*}

\textbf{Qualitative Comparison.} 
Figure~\ref{fig:main_vis} illustrates Q-Zoom's effectiveness in challenging scenarios where critical evidence is tiny and easily destroyed by down-sampling. In the left example (TextVQA), the baseline Qwen2.5-VL-7B hallucinates Pittsburgh due to compression. Conversely, our SD-RPN predicts a concentrated heatmap over the microscopic text, allowing Q-Zoom to accurately read Philadelphia. Similarly, in the right example (V* Bench), the baseline blindly guesses the obscured broom's color as Gray. Q-Zoom seamlessly localizes the object, routing the high-resolution crop to correctly identify it as ``Black.'' These visualizations prove Q-Zoom robustly rescues MLLMs from resolution-induced hallucinations.

\begin{figure*}[t]
	\centering
	\begin{subfigure}[b]{0.48\linewidth}
		\centering
		\includegraphics[width=\linewidth]{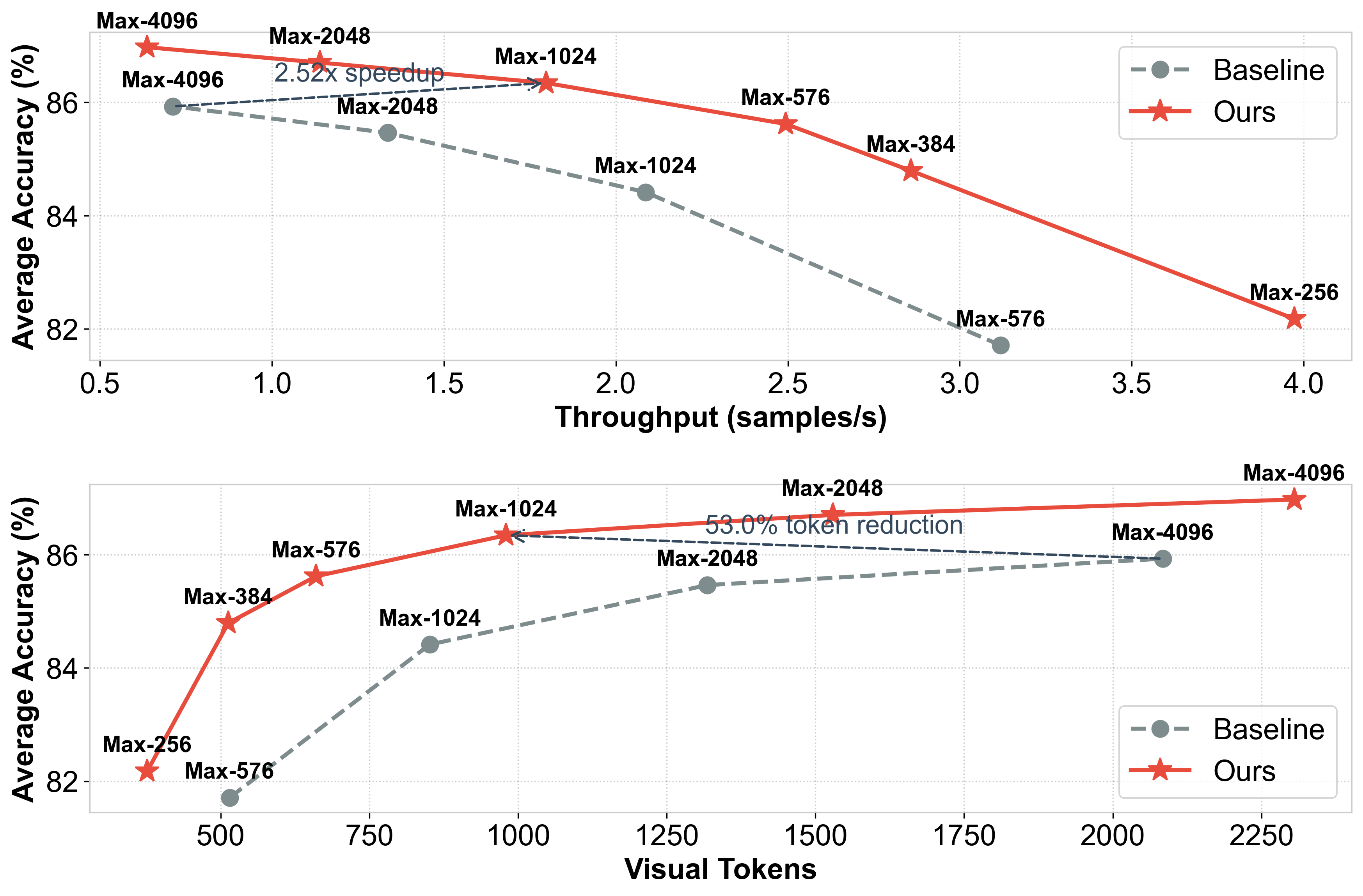} 
		\caption{Document \& OCR Benchmarks}
		\label{fig:curve_doc_ocr}
	\end{subfigure}
	\hfill
	\begin{subfigure}[b]{0.48\linewidth}
		\centering
		\includegraphics[width=\linewidth]{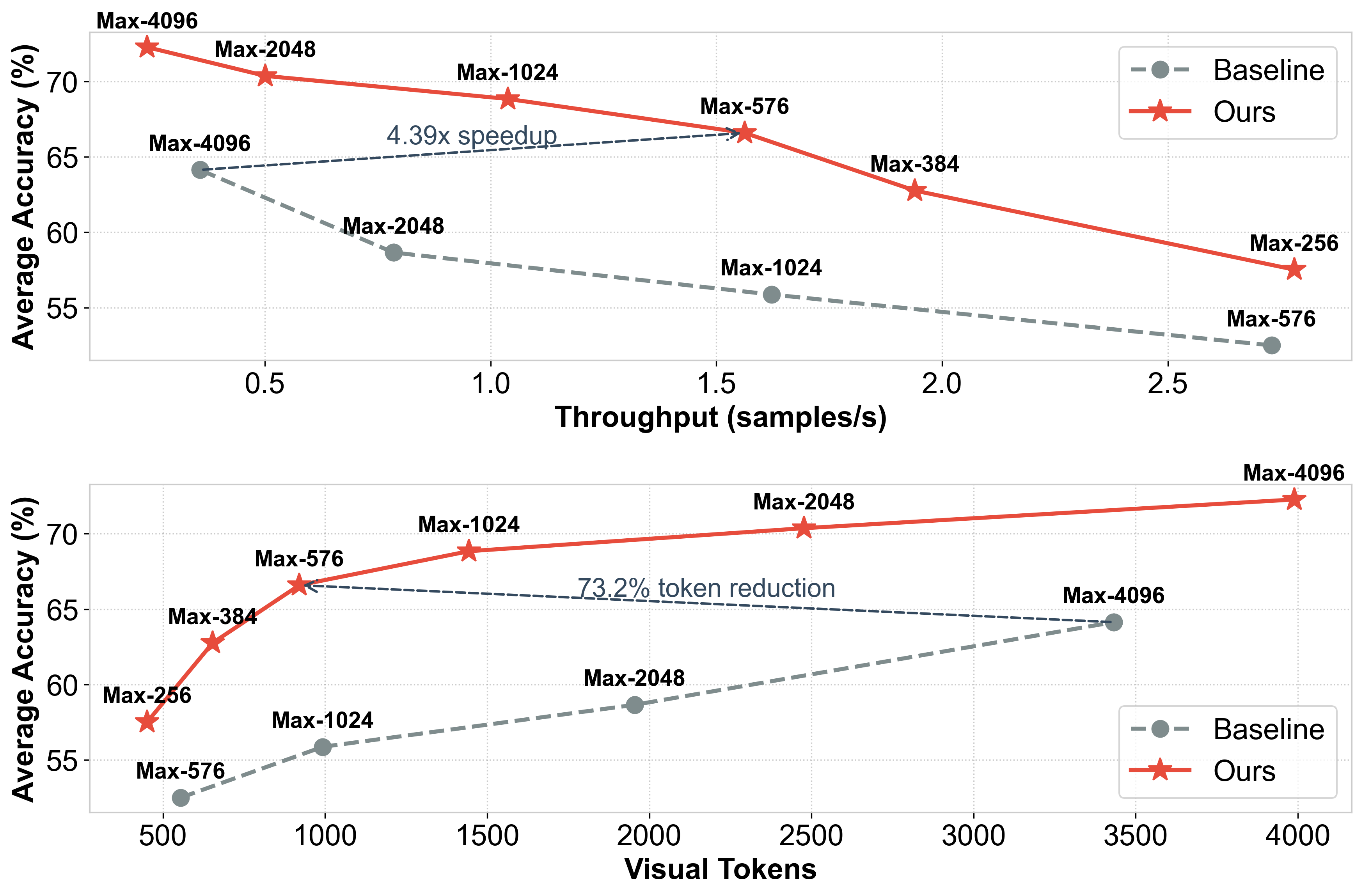} 
		\caption{High-Res \& Vision-Centric Benchmarks}
		\label{fig:curve_hr}
	\end{subfigure}
	
	\vspace{0mm}
	\caption{\textbf{Accuracy vs. Efficiency Trade-offs.} We evaluate the Qwen2.5-VL-7B baseline against Q-Zoom by sweeping the maximum visual token limit in a image from 256 to 4,096. Our framework establishes a dominant Pareto frontier on both \textbf{(a)} Document \& OCR and \textbf{(b)} High-Resolution benchmark categories. By adaptively localizing RoIs, Q-Zoom surpasses the peak accuracy of the brute-force 4,096-token baseline while reducing the visual token cost and accelerating throughput.}
	\label{fig:acc_efficiency_curve}
	\vspace{-2mm}
\end{figure*}

\textbf{Accuracy-Efficiency Trade-off Analysis.} 
To rigorously prove our framework circumvents the quadratic scaling bottleneck, Figure~\ref{fig:acc_efficiency_curve} plots performance-efficiency Pareto frontiers by varying the maximum visual token limit on Qwen2.5-VL-7B. 
On Document \& OCR tasks (Fig.~\ref{fig:acc_efficiency_curve}a), the baseline maxes out at 85.9\% using 4,096 tokens. Q-Zoom surpasses this peak using a maximum of only 1,024 tokens. By adaptively extracting RoIs and bypassing redundant backgrounds, it achieves a \textbf{2.52$\times$ speedup} and a \textbf{53.0\% token reduction} compared to the 4,096-token baseline.
This efficiency gap widens on High-Resolution benchmarks (Fig.~\ref{fig:acc_efficiency_curve}b). The baseline's accuracy degrades rapidly under token constraints, peaking at 64.2\% (4,096 tokens). In contrast, Q-Zoom achieves 66.7\% accuracy using a 576-token maximum, outperforming the baseline's best configuration by 2.5\% while delivering a massive \textbf{4.39$\times$ acceleration} and a \textbf{73.2\% token reduction}. These curves  prove Q-Zoom establishes a dominant Pareto frontier.

\subsection{Ablation Study}
\label{sec:ablation}

\begin{table*}[t]
	\centering
	\setlength{\tabcolsep}{3.5pt}
	\caption{Ablation study on the core components of our proposed framework. We progressively enable the upgraded Self-Distilled Region Proposal Network (\textbf{RPN}), targeted Supervised Fine-Tuning (\textbf{SFT}), and the Dynamic Gating Network (\textbf{Gate}). The $\dagger$ denotes results obtained using the configuration from our preliminary conference version. Tp denotes throughput, which is reported as relative speed ($\times$) to baseline.}
	\label{tab:ablation_components}
	\begin{tabular}{@{} ccc ccccccc cccccc ccc @{} }
		\toprule
		\multicolumn{3}{c}{\textbf{Components}} & \multicolumn{7}{c}{\textbf{Document \& OCR}} & \multicolumn{6}{c}{\textbf{High-Res \& Vision-Centric}} & \multicolumn{3}{c}{\textbf{General QA}} \\
		\cmidrule(lr){1-3} \cmidrule(lr){4-10} \cmidrule(lr){11-16} \cmidrule(lr){17-19}
		\textbf{RPN} & \textbf{SFT} & \textbf{Gate} & \textbf{Tp} & \textbf{Doc} & \textbf{Chart} & \textbf{OCR} & \textbf{Info} & \textbf{Text} & \textbf{Ave.} & \textbf{Tp} & \textbf{V*} & \textbf{RW} & \textbf{HR4K} & \textbf{HR8K} & \textbf{Ave.} & \textbf{Tp} & \textbf{MME} & \textbf{MMS} \\
		\midrule
		
		\rowcolor{black!10!white}
		\multicolumn{19}{l}{\textcolor{black}{\textbf{Qwen2.5-VL-7B}}} \\
		
		- & - & - &
		1.0$\times$ & 92.0 & 83.0 & 82.8 & 70.1 & 81.1 & 81.8 &
		1.0$\times$ & 64.4 & 35.4 & 57.9 & 52.4 & 52.5 &
		1.0$\times$ & 2297 & 62.4 \\
		
		\color{gray}\checkmark$^\dagger$ & \color{gray}- & \color{gray}- &
		\color{gray}0.50$\times$ & \color{gray}93.6 & \color{gray}85.5 & \color{gray}82.9 & \color{gray}76.9 & \color{gray}83.5 & \color{gray}84.5 &
		\color{gray}0.47$\times$ & \color{gray}77.5 & \color{gray}40.5 & \color{gray}73.3 & \color{gray}66.1 & \color{gray}64.4 &
		\color{gray}0.40$\times$ & \color{gray}2335 & \color{gray}62.1 \\
		
		\checkmark & - & - &
		0.59$\times$ & 94.1 & 85.8 & 84.9 & 79.6 & 83.0 & 85.5 &
		0.49$\times$ & 80.1 & 44.6 & 75.5 & 66.6 & 66.7 &
		0.63$\times$ & 2272 & 62.5 \\
		
		\checkmark & \checkmark & - &
		0.63$\times$ & 94.5 & 86.5 & 85.9 & 79.9 & 83.8 & 86.1 &
		0.52$\times$ & 80.1 & 46.0 & 75.8 & 67.4 & 67.3 &
		0.66$\times$ & 2322 & 63.1 \\
		
		\rowcolor{headergray}
		\checkmark & \checkmark & \checkmark &
		0.81$\times$ & 94.3 & 85.6 & 85.4 & 79.4 & 83.5 & 85.6 &
		0.54$\times$ & 79.6 & 45.7 & 74.9 & 66.3 & 66.6 &
		0.84$\times$ & 2326 & 63.4 \\
		
		\midrule
		\rowcolor{black!10!white}
		\multicolumn{19}{l}{\textcolor{black}{\textbf{Qwen3-VL-4B}}} \\
		
		- & - & - &
		1.0$\times$ & 91.3 & 84.0 & 83.1 & 68.0 & 79.2 & 81.1 &
		1.0$\times$ & 62.3 & 40.3 & 62.4 & 56.3 & 55.3 &
		1.0$\times$ & 2335 & 62.8 \\
		
		\checkmark & - & - &
		0.61$\times$ & 92.8 & 84.0 & 84.2 & 75.4 & 78.2 & 82.9 &
		0.56$\times$ & 82.7 & 45.3 & 74.0 & 66.4 & 67.1 &
		0.65$\times$ & 2299 & 61.6 \\
		
		\checkmark & \checkmark & - &
		0.63$\times$ & 93.5 & 85.0 & 84.5 & 77.3 & 81.4 & 84.3 &
		0.55$\times$ & 83.7 & 49.8 & 77.3 & 70.0 & 70.2 &
		0.65$\times$ & 2344 & 63.1 \\
		
		\rowcolor{headergray}
		\checkmark & \checkmark & \checkmark &
		0.82$\times$ & 93.4 & 85.0 & 84.6 & 77.1 & 81.4 & 84.3 &
		0.61$\times$ & 80.1 & 49.9 & 76.5 & 68.5 & 68.8 &
		0.80$\times$ & 2349 & 63.1 \\
		\bottomrule
	\end{tabular}
\end{table*}

\begin{table*}[t]
	\centering
	\setlength{\tabcolsep}{3.5pt}
	\caption{Ablation of training-free pseudo-label based ROI strategies. Throughput is reported as relative speed ($\times$) to each model's baseline under the same macro-category.}
	\label{tab:ablation_pseudo_labels}
	\begin{tabular}{@{}l c ccccc c c cccc c@{}}
		\toprule
		\multirow{2}{*}{\textbf{Method}} & \multicolumn{7}{c}{\textbf{Document \& OCR}} & \multicolumn{6}{c}{\textbf{High-Res \& Vision-Centric}} \\
		\cmidrule(lr){2-8} \cmidrule(lr){9-14}
		& \textbf{Tp} & \textbf{Doc} & \textbf{Chart} & \textbf{OCR} & \textbf{Info} & \textbf{Text} & \textbf{Ave.} & \textbf{Tp} & \textbf{V*} & \textbf{RW} & \textbf{HR4K} & \textbf{HR8K} & \textbf{Ave.} \\
		\midrule
		(0) \textbf{LLaVA-1.5-7B} &
		1.0$\times$ & 21.5 & 18.1 & 31.4 & 20.4 & 46.1 & 27.5 &
		1.0$\times$ & 50.3 & 27.7 & 37.5 & 33.8 & 37.3 \\
		
		(1) Response-to-Image Attention & 
		0.39$\times$ & 28.4 & 22.8 & 32.4 & 22.2 & 53.1 & 31.8 & 
		0.38$\times$ & 57.1 & 27.6 & 41.8 & 35.3 & 40.5 \\
		
		(2) GroundingDINO (1 bbox) & 
		0.37$\times$ & 23.2 & 19.2 & 32.5 & 20.6 & 52.6 & 29.6 & 
		0.17$\times$ & 56.5 & 27.9 & 48.9 & 44.8 & 44.5 \\
		
		(3) GroundingDINO (2 bbox) & 
		0.32$\times$ & 23.8 & 19.2 & 32.5 & 20.2 & 54.5 & 30.0 & 
		0.15$\times$ & 62.3 & 27.6 & 48.0 & 46.3 & 46.1 \\
		
		\rowcolor{headergray}
		(4) +SD-RPN &
		0.62$\times$ & 34.2 & 20.6 & 37.3 & 22.3 & 58.8 & 34.6 & 
		0.57$\times$ & 70.7 & 27.7 & 47.3 & 41.6 & 46.8 \\
		\midrule
		
		(0) \textbf{Qwen2.5-VL-7B} &
		1.0$\times$ & 92.0 & 83.0 & 82.8 & 70.1 & 81.1 & 81.8 &
		1.0$\times$ & 64.4 & 35.4 & 57.9 & 52.4 & 52.5 \\
		
		(1) Response-to-Image Attention & 
		0.34$\times$ & 93.0 & 85.2 & 83.7 & 76.5 & 82.6 & 84.2 & 
		0.25$\times$ & 66.5 & 38.4 & 61.9 & 57.4 & 56.1 \\
		
		(2) GroundingDINO (1 bbox) & 
		0.43$\times$ & 92.6 & 84.4 & 83.0 & 70.8 & 81.5 & 82.5 & 
		0.31$\times$ & 67.5 & 38.6 & 70.1 & 61.7 & 59.5 \\
		
		(3) GroundingDINO (2 bbox) & 
		0.36$\times$ & 92.5 & 84.4 & 82.4 & 70.7 & 81.6 & 82.3 & 
		0.25$\times$ & 66.0 & 40.3 & 70.6 & 63.0 & 60.0 \\
		
		\rowcolor{headergray}
		(4) +SD-RPN &
		0.59$\times$ & 94.1 & 85.8 & 84.9 & 79.6 & 83.0 & 85.5 &
		0.49$\times$ & 80.1 & 44.6 & 75.5 & 66.6 & 66.7 \\
		\bottomrule
	\end{tabular}
\end{table*}

In this subsection, we conduct a comprehensive ablation study to validate our overall framework architecture, the micro-designs of individual modules, and our hyper-parameter selections. For brevity within the tables, benchmarks are abbreviated as follows: Doc (DocVQA), Chart (ChartQA), Info (InfoVQA), Text (TextVQA), RW (MME-RealWorld), HR4K/HR8K (HR-Bench 4K/8K), and MMS (MMStar). Unless otherwise specified, we constrain the maximum visual token limit to 576 across all benchmarks to maintain consistency and strict alignment with our training configurations.

\textbf{Effectiveness of Key Components.} 
We systematically evaluate the contributions of our three primary framework upgrades in Table~\ref{tab:ablation_components}: the SD-RPN, Spatio-Temporal Alignment via targeted Supervised Fine-Tuning (SFT), and the Dynamic Gating Network. Compared to our preliminary version ($\dagger$), removing strict token constraints and enriching the SD-RPN training pool with 33K high-resolution DocVQA samples improves both perceptual accuracy and inference throughput. Integrating targeted SFT explicitly resolves spatial misalignment between dense local RoI tokens and coarse global image tokens. This mitigates contextual distraction, restoring global spatial reasoning without degrading foundational intelligence on General QA benchmarks. Finally, the Dynamic Gating Network successfully balances accuracy and computational cost. On Document/OCR and General QA tasks, it safely bypasses the RoI branch for simpler queries, boosting relative throughput by nearly 30\%. Conversely, on detail-heavy High-Resolution benchmarks, the gate consistently triggers the RoI branch, maintaining peak perceptual accuracy. This dynamic behavior confirms the gate's ability to reliably assess task complexity and allocate resources only where visually necessary.

\textbf{Comparison with Training-Free RoI Strategies.} 
To validate the necessity of a dedicated region proposal module, Table~\ref{tab:ablation_pseudo_labels} compares SD-RPN against two training-free RoI alternatives: (1) thresholding raw Response-to-Image cross-attention maps, and (2) using GroundingDINO~\cite{liu2024grounding} (retaining the top 1 or 2 bounding boxes). The Attention strategy attempts to bypass training but suffers from noisy localization due to irrelevant background artifacts. Furthermore, it severely degrades throughput, as extracting the attention map requires a costly auto-regressive decoding pass before cropping. Alternatively, GroundingDINO lacks deep semantic reasoning, struggling profoundly with complex queries (yielding only marginal gains on Document \& OCR tasks). While it moderately boosts the weaker LLaVA baseline on High-Resolution tasks, it fails to generalize synergistically with stronger models like Qwen2.5-VL. Finally, decoupling the external tool from the MLLM prevents shared computation, resulting in the poorest efficiency. SD-RPN overcomes these issues by directly distilling query-conditioned reasoning into a lightweight, integrated branch, achieving superior accuracy and speed.

%
%
\begin{table}[t]
	\centering
	\setlength{\tabcolsep}{4pt}
	\caption{Ablation on the backbone depth ($B$) and the number of RPN layers ($R$) using Qwen2.5-VL 7B baseline. When ablating $B$, $R$ is fixed to 3; when ablating $R$, $B$ is fixed to 18.}
	\label{tab:ablation_br_layers}
	\begin{tabular}{@{}lc ccccc ccc@{}}
		\toprule
		& \textbf{Value} & \textbf{Doc} & \textbf{Chart} & \textbf{OCR} & \textbf{Info} & \textbf{Text} & \textbf{V*} & \textbf{RW} & \textbf{Ave.} \\
		\midrule
		
		& 3  & 92.6 & 85.0 & 82.2 & 70.4 & 81.2 & 61.8 & 36.5 & 72.8 \\
		& 9  & 93.0 & 84.8 & 83.3 & 71.0 & 81.4 & 71.2 & 36.0 & 74.4 \\
		& 15 & 94.0 & 85.6 & 84.4 & 77.6 & 82.7 & 71.7 & 39.6 & 76.5 \\
		
		\rowcolor{headergray}\cellcolor{white}
		& 18 & \textbf{94.1} & \textbf{85.8} & \textbf{84.9} & \textbf{79.6} & \textbf{83.0} & \textbf{80.1} & \textbf{44.6} & \textbf{78.9} \\
		\multirow{-5}{*}{\textbf{$B$}} & 21 & 93.6 & 85.7 & 84.4 & 76.4 & 82.2 & 78.5 & 41.2 & 77.4 \\
		
		\midrule
		
		& 1 & 93.6 & 85.2 & 84.6 & 75.5 & 82.5 & 71.7 & 40.4 & 76.2 \\
		& 2 & 93.8 & \textbf{85.8} & 84.8 & 78.7 & \textbf{83.1} & 78.6 & 44.2 & 78.4 \\
		
		\rowcolor{headergray}\cellcolor{white}
		& 3 & \textbf{94.1} & \textbf{85.8} & \textbf{84.9} & \textbf{79.6} & 83.0 & \textbf{80.1} & \textbf{44.6} & \textbf{78.9} \\
		\multirow{-4}{*}{\textbf{$R$}} & 4 & 93.6 & \textbf{85.8} & 84.1 & 76.9 & 82.4 & 75.4 & 44.5 & 77.5 \\
		\bottomrule
	\end{tabular}
\end{table}

\textbf{Impact of Backbone Depth and SD-RPN Capacity.} 
Table~\ref{tab:ablation_br_layers} evaluates the optimal backbone split depth ($B$) and tunable RPN transformer layers ($R$) using the Qwen2.5-VL-7B baseline. Fixing $R=3$, we sweep $B$ from layer 3 to 21. Perceptual accuracy improves progressively, peaking at $B=18$ before degrading. This empirically optimal depth aligns exactly with the inherent localization layers identified in recent probing studies~\cite{shi2025vision}. Next, fixing $B=18$, we ablate $R$ from 1 to 4. A single-layer projection yields suboptimal localization, showing the network requires sufficient depth to translate intermediate features into dense heatmaps. Performance strictly improves up to $R=3$, with a slight regression at $R=4$. Consequently, we adopt $B=18$ and $R=3$ across all main experiments to guarantee optimal efficiency and precision.
%
\begin{table}[t]
	\centering
	\setlength{\tabcolsep}{4pt}
	\caption{Ablation on pseudo-label training data size for the SD-RPN. The $\dagger$ denotes a baseline model supervised exclusively by 68K ground-truth (GT) bounding boxes from the Visual CoT dataset.}
	\label{tab:ablation_data_size}
	\begin{tabular}{@{}c ccccc ccc@{}}
		\toprule
		\textbf{Data Size} & \textbf{Doc} & \textbf{Chart} & \textbf{OCR} & \textbf{Info} & \textbf{Text} & \textbf{V*} & \textbf{RW} & \textbf{Ave.} \\
		\midrule
		10k  & 93.6 & 85.4 & 84.9 & 78.0 & 82.7 & 78.5 & 40.7 & 77.7 \\
		25k  & 93.5 & 85.5 & \textbf{85.1} & 78.1 & \textbf{83.0} & \textbf{80.1} & 42.4 & 78.2 \\
		50k  & 93.4 & 85.5 & 84.6 & 78.8 & \textbf{83.0} & 78.0 & 43.7 & 78.1 \\
		100k & 93.6 & 85.3 & 84.9 & 79.2 & 82.9 & 78.0 & \textbf{45.1} & 78.4 \\
		\rowcolor{headergray}
		185k & \textbf{94.1} & \textbf{85.8} & 84.9 & \textbf{79.6} & \textbf{83.0} & \textbf{80.1} & 44.6 & \textbf{78.9} \\
		\midrule
		68k$^\dagger$ & 93.4 & 85.5 & 83.6 & 75.8 & \textbf{83.0} & \textbf{80.1} & 44.3 & 78.0 \\
		\bottomrule
	\end{tabular}
\end{table}

\textbf{Data Efficiency and Self-Distillation.} 
Table~\ref{tab:ablation_data_size} evaluates SD-RPN's data scalability and pseudo-label quality. The module demonstrates exceptionally fast convergence: fine-tuning with only 10K self-distilled pseudo-labels yields robust perceptual enhancement (77.7\% average). Scaling to the full 185K dataset ensures a steady performance trajectory, peaking at 78.9\%. To explicitly validate the efficacy of our training-free label generation, we establish a GT supervised baseline. For this variant (denoted by $\dagger$), we bypassed the pseudo-label generation entirely and trained the SD-RPN using 68K GT bounding boxes sampled from the Visual CoT training set~\cite{shao2024visual} (comprising 50K samples from GQA and 18K samples from TextVQA). This GT-supervised model averages 78.0\%, directly comparable to our self-distilled model trained on just 50K pseudo-labels (78.1\%). This confirms our distillation pipeline successfully eliminates dependency on external annotated datasets without compromising performance.
%
\begin{table}[t]
	\centering
	\caption{Ablation study on pseudo-label thresholds $\tau_{fg}$ and $\tau_{bg}$ across different model architectures.}
	\label{tab:threshold_ablation}
	\setlength{\tabcolsep}{4pt} 
	\begin{tabular}{@{}ll cccc cccc@{}}
		\toprule
		\multirow{2}{*}{$\boldsymbol{\tau_{fg}}$} & \multirow{2}{*}{$\boldsymbol{\tau_{bg}}$} & \multicolumn{4}{c}{\textbf{LLaVA-1.5-7B}} & \multicolumn{4}{c}{\textbf{Qwen2.5-VL-7B}} \\
		\cmidrule(lr){3-6} \cmidrule(lr){7-10}
		& & \textbf{OCR} & \textbf{Text} & \textbf{V*} & \textbf{Ave.} & \textbf{OCR} & \textbf{Text} & \textbf{V*} & \textbf{Ave.} \\
		\midrule
		\multicolumn{2}{@{}l}{Baseline} & 31.4 & 46.1 & 50.3 & 42.6 & 82.8 & 81.1 & 64.4 & 76.1 \\
		\midrule
		0.10 & 0.10 & 35.1 & 57.4 & 65.4 & 52.6 & 84.3 & \textbf{83.1} & 78.0 & 81.8 \\
		0.15 & 0.10 & 36.3 & 57.9 & 67.5 & 53.9 & 84.2 & \textbf{83.1} & 79.6 & 82.3 \\
		0.20 & 0.10 & 37.0 & 58.7 & 67.5 & 54.4 & 84.5 & 82.8 & 79.1 & 82.1 \\
		0.30 & 0.10 & 36.6 & 58.1 & 64.9 & 53.2 & 84.2 & 82.6 & 77.5 & 81.4 \\
		\rowcolor{headergray}
		\textbf{0.20} & \textbf{0.05} & 37.3 & \textbf{58.8} & \textbf{70.7} & \textbf{55.6} & \textbf{84.9} & 83.0 & \textbf{80.1} & \textbf{82.7} \\
		0.20 & 0.03 & \textbf{37.8} & 58.7 & 68.6 & 55.0 & 84.8 & 83.0 & 79.1 & 82.3 \\
		\bottomrule
	\end{tabular}
\end{table}

\textbf{Impact of Pseudo-Label Assignment Thresholds.} 
Table~\ref{tab:threshold_ablation} ablates the pseudo-label foreground ($\tau_{fg}$) and background ($\tau_{bg}$) thresholds.
Setting the two thresholds to the same value (e.g., $\tau_{fg} = \tau_{bg} = 0.10$) forces a hard, naive binary classification over the raw attention maps. This configuration effectively disables our proposed tri-state label assignment design, forcing the network to train on the highly ambiguous middle-range tokens. This leads to a noticeable performance downgrade. 
Sweeping these boundaries reveals that with $\tau_{bg}=0.10$, a moderate foreground margin ($\tau_{fg}=0.20$) maximizes precision. Coupling this optimal boundary with an aggressive background filter ($\tau_{fg}=0.20, \tau_{bg}=0.05$) yields peak average performance across architectures, perfectly balancing distillation purity with structural RoI coverage.
	
	
	

\begin{figure}[t]
	\centering
	
	\includegraphics[width=0.85\linewidth]{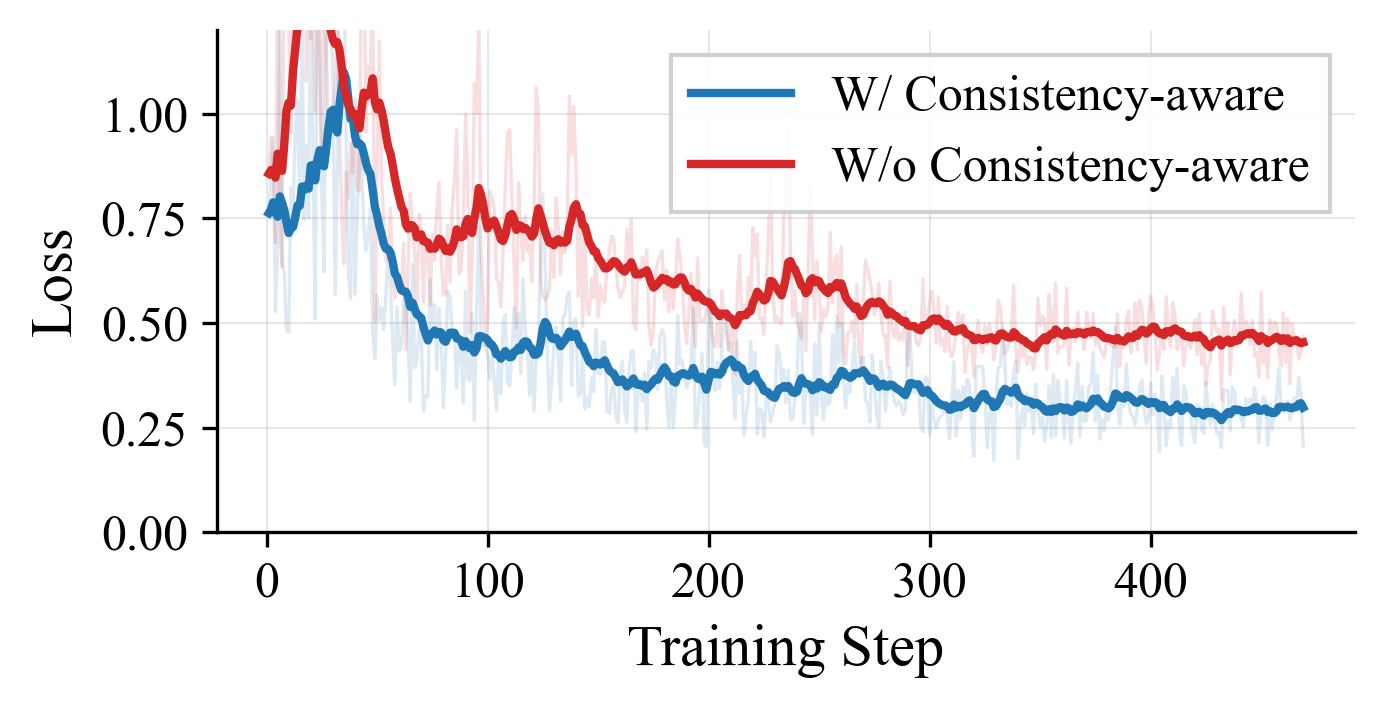}
	
	\vspace{0mm} 
	
	\includegraphics[width=0.85\linewidth]{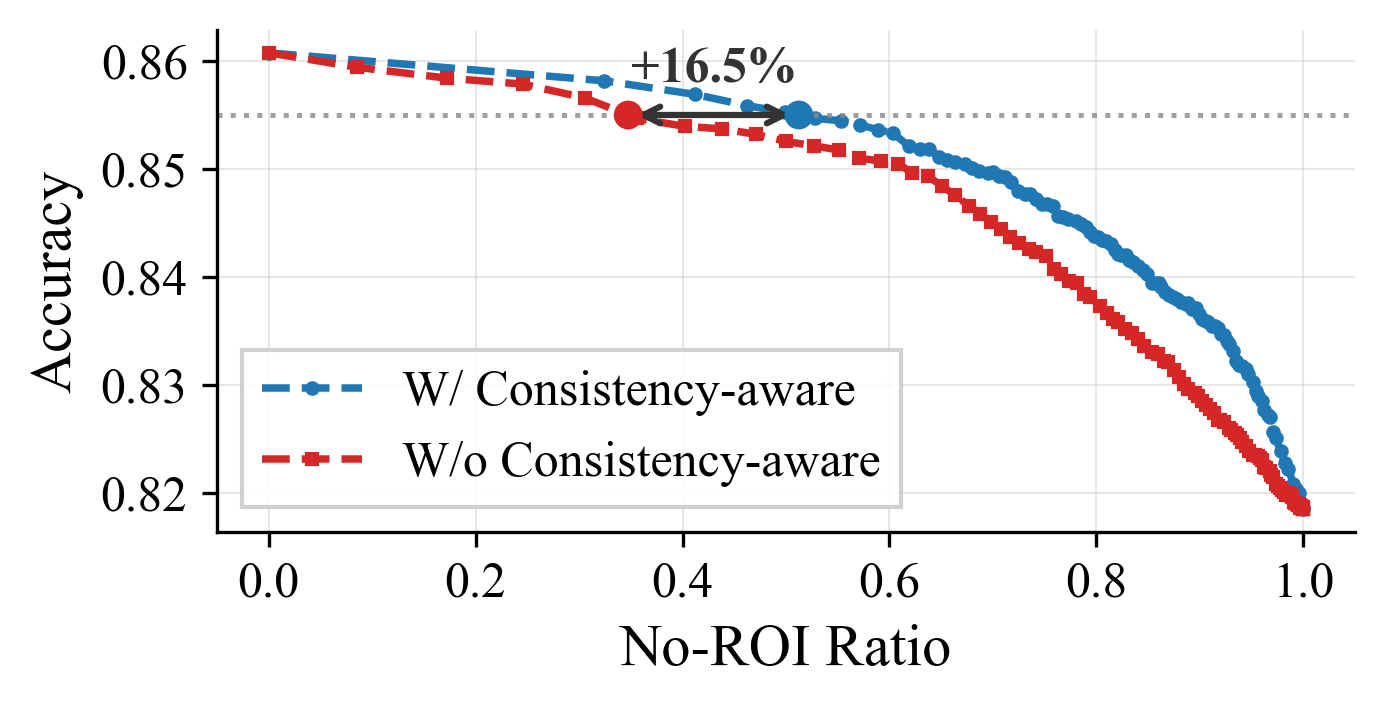}
	
	\vspace{0mm}
	\caption{\textbf{Ablation of the Consistency-aware Training Sample Generation upon Qwen2.5-VL 7B.} \textbf{(Top)} Training loss curves of the dynamic gating network. \textbf{(Bottom)} The Pareto front illustrating the trade-off between perception accuracy and inference efficiency (No-RoI Ratio).}
	\label{fig:ablation_consistency}
	\vspace{-4mm}
\end{figure}

\textbf{Effectiveness of Consistency-aware Sample Generation.} 
To empirically validate our Consistency-aware Training Sample Generation, Figure~\ref{fig:ablation_consistency}(a) compares it against a naive labeling baseline (Section~\ref{sec:label_construction}) on Qwen2.5-VL 7B. Training the gating network with naive, noise-corrupted labels severely destabilizes optimization, plateauing at a high loss bound. Enforcing a multi-resolution consistency check smooths optimization, converging faster to a lower bound, which directly translates to more robust inference-time routing. Figure~\ref{fig:ablation_consistency}(b) plots the Pareto frontier of overall accuracy against the No-RoI Ratio (queries successfully routed to the low-resolution pathway) across five Doc and OCR benchmarks. Our consistency-aware gate exhibits strict Pareto dominance over the naive baseline. At an 85.5\% accuracy threshold, it safely bypasses RoI extraction for an additional 16.5\% of user queries compared to the baseline, improving overall throughput without sacrificing perceptual fidelity.

\section{Conclusion}
In this paper, we presented \textbf{Q-Zoom}, an efficient, query-aware adaptive high-resolution perception framework for MLLMs. Current global resolution scaling paradigms suffer from profound query-level and spatial redundancies, indiscriminately flooding self-attention mechanisms with visually useless tokens. To resolve this, Q-Zoom fundamentally decouples perceptual fidelity from computational cost by dynamically determining \textit{if} high-resolution refinement is necessary and \textit{where} it should be spatially applied.
At its core, Q-Zoom utilizes two lightweight modules operating on the intermediate feature space during the initial prefilling stage. First, the Dynamic Gating Mechanism, optimized via a consistency-aware sample generation strategy, acts as an intelligent router that bypasses high-resolution processing for simpler queries. Second, the Self-Distilled Region Proposal Network (SD-RPN) precisely localizes task-relevant visual evidence for detail-demanding tasks. By employing a fully self-supervised tri-state distillation paradigm, SD-RPN achieves exceptional data efficiency without human annotations, external detection experts, or computationally expensive reinforcement learning. Furthermore, we resolve the inherent perceptual disconnect between cropped regions and the global context using a continuous spatio-temporal positional encoding scheme coupled with targeted Post-SFT, fully restoring the model's spatial reasoning capabilities. 
Extensive evaluations across Document, OCR, and High-Resolution benchmarks conclusively prove that Q-Zoom establishes a dominant Pareto frontier, offering a robust, scalable, and highly accessible paradigm for efficient visual perception in MLLMs.
	
	\bibliographystyle{IEEEtran}
	\bibliography{references}

	\clearpage
	\appendices
\section{Implementation and Prompt Details} 
\label{app:implementation}

\subsection{More Implementation Details}
\label{app:more_impl_details}

\textbf{Training Configurations.} The optimization hyperparameters for the three core components of Q-Zoom are detailed in Table~\ref{tab:qzoom_stage_settings}. Across all components, we universally apply the AdamW optimizer with a weight decay of 0.0, momentum parameters $(\beta_1, \beta_2) = (0.9, 0.98)$, and a cosine learning rate decay scheduler with a 3\% linear warmup ratio. Each component is trained for a single epoch.

\begin{table}[hbt!]
	\centering
	\caption{Training hyperparameters for the three core components of Q-Zoom.}
	\label{tab:qzoom_stage_settings}
	\resizebox{\columnwidth}{!}{%
		\begin{tabular}{@{}lccc@{}}
			\toprule
			\textbf{Config} & \textbf{SD-RPN} & \textbf{Post-SFT} & \textbf{Dynamic Gate} \\
			\midrule
			Optimizer & \multicolumn{3}{c}{AdamW} \\
			Weight decay & \multicolumn{3}{c}{0.0} \\
			Optimizer momentum & \multicolumn{3}{c}{$(\beta_1, \beta_2) = (0.9, 0.98)$} \\
			Batch size & 128 & 64 & 128 \\
			Learning rate schedule & \multicolumn{3}{c}{cosine decay} \\
			Peak learning rate & $1\times10^{-4}$ & $1\times10^{-6}$ & $1\times10^{-4}$ \\
			Warm-up strategy & \multicolumn{3}{c}{linear} \\
			Warm-up ratio & \multicolumn{3}{c}{0.03} \\
			Max gradient norm & \multicolumn{3}{c}{1.0} \\
			Training epochs & \multicolumn{3}{c}{1} \\
			\bottomrule
		\end{tabular}%
	}
\end{table}

\textbf{Dataset Usage and Filtering.} 
Table~\ref{tab:qzoom_dataset_usage} provides a comprehensive breakdown of the training datasets and sample sizes utilized across our core components. To ensure robust generalization, our base data mixture spans standard visual question answering (e.g., GQA~\cite{hudson2019gqa} and OCR-VQA~\cite{mishra2019ocr} from LLaVA-1.5~\cite{liu2023improvedllava}), document and chart understanding (e.g., DocVQA~\cite{mathew2020docvqa} and ChartQA~\cite{masry2022chartqa} from Visual CoT~\cite{shao2024visual}), and fine-grained spatial reasoning (e.g., V*-COCO~\cite{vstar}). Importantly, for the Post-SFT and Dynamic Gate stages, we do not utilize these raw datasets in their entirety. Instead, as detailed in the main text, we apply rigorous selective filtering to construct highly targeted training sets. Specifically, we mine contrastive hard-regression cases to resolve contextual distraction during the Post-SFT stage (yielding $\sim$7K samples), and isolate valid multi-resolution routing behaviors for the Dynamic Gate (yielding 40K--60K samples, depending on the inherent performance of the specific base model).

\begin{table}[hbt!]
	\centering
	\caption{Training data used by each core component of Q-Zoom.}
	\label{tab:qzoom_dataset_usage}
	\resizebox{\columnwidth}{!}{%
		\begin{tabular}{@{}lllr@{}}
			\toprule
			\textbf{Component} & \textbf{Model} & \textbf{Training Source} & \textbf{Samples} \\
			\midrule
			\multirow{7}{*}{\textbf{SD-RPN}} 
			& \multirow{4}{*}{Qwen-series} 
			& GQA & 72K \\
			& & OCR-VQA & 80K \\
			& & VCoT-DocVQA & 33K \\
			\cmidrule(l){3-4}
			& & \textit{Total} & \textit{185K} \\
			\cmidrule(l){2-4}
			& \multirow{3}{*}{LLaVA-series} 
			& GQA & 72K \\
			& & OCR-VQA & 80K \\
			\cmidrule(l){3-4}
			& & \textit{Total} & \textit{152K} \\
			\midrule
			\multirow{6}{*}{\textbf{Post-SFT}}
			& \multirow{6}{*}{Qwen-series} 
			& TextVQA$_{\text{train}}$ & 34K \\
			& & ChartQA$_{\text{train}}$ & 28K \\
			& & VCoT-InfoVQA & 15K \\
			& & VCoT-DocVQA & 33K \\
			& & V*-COCO & 44K \\
			\cmidrule(l){3-4}
			& & \textit{Mined Hard Samples} & \textit{$\sim$7K} \\
			\midrule
			\multirow{5}{*}{\textbf{Dynamic Gate}}
			& \multirow{5}{*}{All-models} 
			& VCoT-TextVQA & 18K \\
			& & VCoT-GQA & 50K \\
			& & VCoT-DocVQA & 33K \\
			& & ChartQA$_{\text{train}}$ & 28K \\
			\cmidrule(l){3-4}
			& & \textit{Filtered Training Set} & \textit{40K--60K} \\
			\bottomrule
		\end{tabular}%
	}
\end{table}

\textbf{Backbone and Branch Configurations.} Table~\ref{tab:qzoom_twig} details the specific backbone split depths ($B$) for all models evaluated in our primary experiments. Across all model variants, we maintain a constant branch depth of $R=3$ for both the SD-RPN and dynamic gating modules. Furthermore, the Zoom-within-Zoom (ZwZ) variants strictly inherit the structural settings of their corresponding base models.

\begin{table}[hbt!]
	\centering
	\caption{Backbone split depth $B$ for the models used in the main results. All models use the same branch depth $R=3$ for both SD-RPN and dynamic gating.}
	\label{tab:qzoom_twig}
	\begin{tabular}{lc}
		\toprule
		\textbf{Model} & \textbf{Backbone Layer $B$} \\
		\midrule
		LLaVA-1.5-7B & 15 \\
		LLaVA-1.5-13B & 15 \\
		Qwen2.5-VL-3B & 24 \\
		Qwen2.5-VL-7B & 18 \\
		ZwZ-Qwen2.5-VL-7B & 18 \\
		Qwen3-VL-4B & 24 \\
		ZwZ-Qwen3-VL-4B & 24 \\
		\bottomrule
	\end{tabular}
\end{table}

\subsection{Prompt Usage}
\label{app:prompt_usage}

\tcbset{
	promptstyle/.style={
		colback=black!5,            
		colframe=black,             
		colbacktitle=black,         
		coltitle=white,             
		fonttitle=\bfseries\ttfamily, 
		fontupper=\small\ttfamily,  
		arc=1.5mm,                  
		boxrule=0.8pt,              
		left=8pt,
		right=8pt,
		top=6pt,
		bottom=6pt,
		toptitle=4pt,               
		bottomtitle=4pt,
	}
}

\textbf{SD-RPN Prompts.} For LLaVA-1.5 and the Qwen-series textual mode, we adopt the standard short-answer prompt format, shown below as the DEFAULT PROMPT. However, we observe that the attention distributions of Qwen-series models differ substantially between textual/document images and natural images. To address this, we introduce a distinct natural-image prompt for the Qwen models, referred to as the QWEN SD-RPN NATURAL MODE prompt. This variant appends an explicit grounding instruction to encourage spatially localized attention.

\begin{tcolorbox}[promptstyle, title=DEFAULT PROMPT]
	<image> USER: \{question\} Answer the question using a single word or phrase.
\end{tcolorbox}

\begin{tcolorbox}[promptstyle, title=QWEN SD-RPN NATURAL MODE]
	\{original\_question\} Output the grounding bounding box of Region of Interest for the question. IMPORTANT: The output MUST be raw text, one box per line. DO NOT use JSON. Follow this exact format: x\_min y\_min x\_max y\_max \{detail\_label\}.
\end{tcolorbox}

\textbf{LLM-as-a-Judge for Post-SFT.} To identify regression cases where the base model succeeds but the RoI model fails, we employ an LLM-as-a-Judge to evaluate both predictions against the ground truth. The exact instructions are detailed in the QWEN POST-SFT JUDGE PROMPT. This procedure is exclusive to the Qwen-series Post-SFT stage. The judge model is always selected from the same model family at an equal or larger scale. Specifically, we use Qwen2.5-VL-7B-Instruct as the judge for Qwen2.5-VL-3B and Qwen2.5-VL-7B, and Qwen3-VL-8B-Instruct as the judge for Qwen3-VL-4B and Qwen3-VL-8B.

\begin{tcolorbox}[promptstyle, title=QWEN POST-SFT JUDGE PROMPT]
	\textbf{[System Prompt]}\\
	You are a fair and robust evaluator. You focus on semantic correctness over strict string matching.\\[1em]
	\textbf{[User Prompt]}\\
	You are an intelligent evaluator for a Visual Question Answering task.\\
	Compare the Model Predictions against the Ground Truth.\\[0.5em]
	\textbf{--- DATA ---}\\
	Question: \{question\}\\
	Ground Truth: \{ground\_truth\}\\
	Model A Prediction: \{base\_response\}\\
	Model B Prediction: \{roi\_response\}\\[0.5em]
	\textbf{--- CRITERIA ---}\\
	1. \textbf{Containment}: If the Ground Truth is a short value (e.g., '6000') and the Model Prediction contains it correctly within a sentence (e.g., 'The value is 6000'), mark it as \textbf{Yes}.\\
	2. \textbf{Synonyms}: Accept standard synonyms (e.g., 'Yes' = 'True', '10\%' = '0.1').\\
	3. \textbf{Formatting}: Ignore punctuation or capitalization differences (e.g., 'april' = 'April').\\
	4. \textbf{Contradiction}: Only mark 'No' if the prediction is factually wrong or missing the key information.\\[0.5em]
	\textbf{--- TASK ---}\\
	Does Model A provide the correct answer? (Yes/No)\\
	Does Model B provide the correct answer? (Yes/No)\\[0.5em]
	Respond in this exact format:\\
	A: [Yes/No]\\
	B: [Yes/No]
\end{tcolorbox}

\end{document}